\documentclass[twocolumn]{article}

\usepackage{hyperref}
\usepackage{placeins}
\usepackage{adjustbox}
\usepackage{sectsty}
\usepackage{helvet}
\usepackage{appendix}

\usepackage[
letterpaper,
			,includeheadfoot,%
			layoutsize={8.125in,10.875in},%
                layouthoffset=0.1875in,%
                layoutvoffset=0.0625in,%
                left=38.5pt,%
                right=43pt,%
                top=43pt,
                bottom=44pt,%
                headheight=0pt,
                headsep=-2pt,%
                columnsep=0.3125in,
                footskip=25pt,
                marginparwidth=38pt]{geometry}
\usepackage{authblk}
\usepackage{booktabs}
\usepackage{array}
\usepackage{graphicx}
\usepackage{multirow}
\usepackage{amsmath}
\DeclareMathOperator*{\mean}{mean}
\DeclareMathOperator*{\std}{std}
\DeclareMathOperator*{\AbsDiff}{AbsDiff}
\usepackage{amssymb} 
\usepackage{float}
\usepackage{subfig}
\usepackage{pdfpages}
\usepackage{adjustbox}
\usepackage{blindtext}
\usepackage[numbers,sort&compress,merge,square]{natbib}
\usepackage{fancyhdr}
\fancypagestyle{firstpage}{%
  \fancyhf{}%
  \fancyfoot[C]{\footnotesize
    Author-accepted version. The final version of the paper is published in the Proceedings of the 2025 ACM Conference on Fairness, Accountability, and Transparency (FAccT ’25):
    \href{https://doi.org/10.1145/3715275.3732041}{https://doi.org/10.1145/3715275.3732041}}%
}

\usepackage{caption}
\let\oldparagraph\paragraph
\renewcommand{\paragraph}[1]{\oldparagraph{#1.}}

\usepackage{amsmath,amsfonts,bm}

\def\eqref#1{equation~\ref{#1}}

\def\1{\bm{1}}

\def\va{{\bm{a}}}

\def\vb{{\bm{b}}}

\def\vd{{\bm{d}}}

\def\vi{{\bm{i}}}

\def\vt{{\bm{t}}}

\def\vz{{\bm{z}}}

\def\mA{{\bm{A}}}

\def\mB{{\bm{B}}}

\def\mD{{\bm{D}}}

\def\mI{{\bm{I}}}

\DeclareMathAlphabet{\mathsfit}{\encodingdefault}{\sfdefault}{m}{sl}

\SetMathAlphabet{\mathsfit}{bold}{\encodingdefault}{\sfdefault}{bx}{n}

\title{\LARGE Social Perception of Faces in a Vision-Language Model}

\date{}

\author[1,2]{Carina I. Hausladen}
\author[1,3]{Manuel Knott}
\author[1]{Colin F. Camerer}
\author[1]{Pietro Perona}

{
\footnotesize
\affil[1]{California Institute of Technology}
\affil[2]{ETH Zurich, Computational Social Science}
\affil[3]{ETH Zurich, Swiss Data Science Center, Empa}
}

\begin{document}

\twocolumn[
  \begin{@twocolumnfalse}
    
\maketitle
\thispagestyle{firstpage}

\begin{abstract}
\noindent
We explore social perception of human faces in CLIP, a widely used open-source vision-language model. 
To this end, we compare the similarity in CLIP embeddings between different textual prompts and a set of face images. 
Our textual prompts are constructed from well-validated social psychology terms denoting social perception. 
The face images are synthetic and are systematically and independently varied along six dimensions: the legally protected attributes of age, gender, and race, as well as facial expression, lighting, and pose. Independently and systematically manipulating face attributes allows us to study the effect of each on social perception and avoids confounds that can occur in wild-collected data due to uncontrolled systematic correlations between attributes. Thus, our findings are experimental rather than observational.
Our main findings are three. First, while CLIP is trained on the widest variety of images and texts, it is able to make fine-grained human-like social judgments on face images. Second, age, gender, and race do systematically impact CLIP's social perception of faces, suggesting an undesirable bias in CLIP vis-a-vis legally protected attributes. Most strikingly, we find a strong pattern of bias concerning the faces of Black women, where CLIP produces extreme values of social perception across different ages and facial expressions. Third, facial expression impacts social perception more than age and lighting as much as age. The last finding predicts that studies that do not control for unprotected visual attributes may reach the wrong conclusions on bias. 
Our novel method of investigation, which is founded on the social psychology literature and on the experiments involving the manipulation of individual attributes, yields sharper and more reliable observations than previous observational methods and may be applied to study biases in any vision-language model.
\vspace{0.50cm}
\end{abstract}
  \end{@twocolumnfalse}
]

\renewcommand{\thefootnote}{\fnsymbol{footnote}}
\footnotetext[0]{Code available at: \url{https://github.com/carinahausladen/clip-face-bias}}
\renewcommand{\thefootnote}{\arabic{footnote}}
\setcounter{footnote}{0}

\section{Introduction}
Vision-language models (VLMs) are machine learning models trained on large bodies of text and images that can be used to accomplish a variety of tasks, including image search, text-to-image generation, and visual classification~\cite{Zhang2023VisionLanguageMF}. They are increasingly considered for application in industry, entertainment, consumer products, and medical diagnosis thanks to a remarkable human-like ability to make {\em broad} associations between texts and images across a wide variety of domains. However, VLMs' ability to carry out {\em fine-grained} judgments is relatively unexplored. 

Humans spontaneously and quickly form opinions about someone's trustworthiness, dominance, sincerity, intelligence, and other traits just by looking at their face in a photograph~\cite{oosterhof2008functional,sutherland2018facial,todorov2017face, lin2021four}. 
Such ``social perceptions'' are not always accurate and yet can influence the outcome of social decision-making \cite{lin2017cultural,lin2018inferring,todorov2005inferences,antonakis2017looking,hamermesh2011beauty, gallo2024perceived}.

Here we investigate whether VLMs may also be able to make social judgments on photographs of faces and, if so, how such judgments may be affected by facial expression and other facial attributes. Understanding socially relevant behaviors of AI systems is necessary to use them responsibly, and this motivates our interest.

We focus on CLIP~\citep{radford_learning_2021}, a state-of-the-art open-source VLM that is widely used for zero-shot classification \citep{radford_learning_2021}, image retrieval \citep{agarwal_evaluating_2021}, and guiding generative text-to-image models \citep{rombach_high-resolution_2022, ramesh_hierarchical_2022, saharia_photorealistic_2022, balaji_ediff-i_2022}. 
CLIP models are trained on large vision-language-pair datasets \cite[e.g.,][]{schuhmann2021laion} gathered from the internet. 

Despite the fact that CLIP's training set covers the broadest set of topics, including landscapes, animals, man-made objects, and food, our study finds that CLIP indeed can make social judgments about faces in pictures.

We next ask whether CLIP's social judgments are related to human ones and, if so, whether such judgments may mirror some of the common human biases. Forms of social bias in CLIP have been previously documented~\citep{agarwal_evaluating_2021, dehouche_implicit_2021, wang_are_2021, wolfe_evidence_2022, geyik_fairness-aware_2019,ali2023evaluating}, as has bias in CLIP-guided generative text-to-image models (e.g., Stable Diffusion)~\citep{lee2023survey, orgad_editing_2023, zhang_auditing_2023, cho_dall-eval_2022, luccioni_stable_2023}. For example, input-output bias has been measured by systematically prompting the model with language (such as ``show me a CEO'') and then analyzing the demographics of the generated or retrieved image output. 

We approach the study of social judgment of face images by comparing its internal representation, commonly called the {\em embedding}, of image and text pairs. We infer the strength of the vision-language association by measuring the cosine similarity between a text and an image CLIP embeddings~\citep{radford_learning_2021}. 

For example, if the text prompt ``CEO'' is more strongly associated (cosine-similar) with images of men than women \citep{openai_reducing_2022,bianchi_easily_2023}, that demonstrates a gender bias. To be clear, ``bias'' here is judged against a statistically neutral standard, not against actual ecological frequencies (as in the US, there are more male than female CEOs in large firms). 

\begin{figure}[t]
    \centering
    \includegraphics[width=\linewidth]{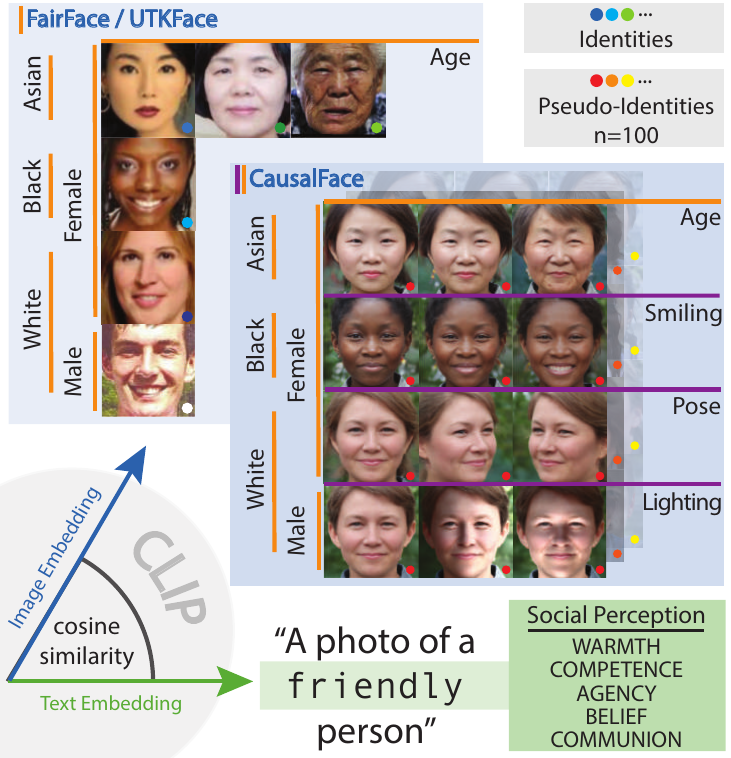}
    \caption{\textbf{Overview of our method.} We study social perception of human faces in CLIP, a vision-language model.  Given one face image and a text prompt including a socially relevant attribute, we infer their CLIP embeddings and compute their cosine similarity within their shared latent space. Images are culled from three datasets: FairFace, UTKFace, CausalFace. FairFace and UTKFace are real-world (wild-collected) datasets which are annotated for legally protected variables, such as gender, race, and age. CausalFace is a synthetic dataset created by experimentally varying both protected (orange) and non-protected (purple) attributes. Text prompts are constructed based on attributes sourced from two theories quantifying social perception in humans (green box, bottom-right). To illustrate, a sub-trait of ``warmth'' is measured by the text prompt of ``a photo of a friendly person'' (friendliness is a warmth-associated word according to the Stereotype Content Model~\citep{fiske_universal_2007}).}
    \vspace{-6pt}
    \label{fig:overview}
\end{figure}

\autoref{fig:overview} summarizes our method, combining two elements in a new way: (1) We use social perception language prompts that are well-validated in social psychology; (2) we use synthetic images to control for spurious correlations with unmeasured visual confounds. Our method yields new observations and higher-quality data. 

Our first innovation is using text prompts that are best at representing social perceptions based on evidence from social psychology. 
Previous studies use text prompts selected ad-hoc by the authors, e.g. denoting occupations or social roles \cite{naik2023social, fraser2024examining, howard2024social}. 
In contrast, we rely on decades of social psychology literature and use words that have emerged as the consensus to represent the major ways in which people categorize each other.
This literature investigates the semantic dimensions that perceivers use to predict the character and intentions of other people.
Some attributes are of greater importance for effectively coordinating social behavior than others and thus serve as fundamental dimensions of social perception \cite{koch_abc_2016, lin2021four}. 
According to the stereotype content model (SCM; \cite{fiske2002model}), the most relevant criteria are the person’s intentions (Warmth) and their ability to carry out their plans (Competence).
A more recent variant of the SCM model subdivided the two traits into three, named ``Communion'' (akin to Warmth), ``Agency'' (socio-economic success, akin to Competence), and a third dimension of political ideology ``Beliefs'' (progressive--conservative)~\citep{koch_abc_2016}. 
We probe CLIP using text prompts expressing the content of both theories to test the robustness of results across them. These psychologically validated stereotype prompts have been used to study bias in AI language algorithms~\citep{cao_large_2023,ungless_robust_2022,fraser_friendly_2023, otterbacher_competent_2017,jeoung2023stereomap}. We introduce them to the study of bias in VLMs. 

A second methodological innovation is adopting an experimental paradigm that allows us to make causal inferences. Multimodal bias is predominantly measured using image datasets that are collected in natural settings and annotated by humans for protected attributes. Unfortunately, as in all observational studies, the distribution of other unlabeled attributes---such as (in the case of face images) lighting, pose, facial expression, and image color statistics---is likely to be correlated with the labeled protected attributes of interest. In other words, unmeasured attributes can create hidden confounds for observational studies. For example, if younger people tend to smile more in photographs, and a smiling face is judged as emotionally warmer, that correlation of smiling with age will confound and inflate the degree to which younger people are judged as warmer. Thus, when bias is found, it is unclear whether bias belongs to the algorithm, to the test data, or both. 
Furthermore, obtaining sufficiently diverse samples across a wide range of intersectional groups can be difficult because some intersections are relatively rare. 
Thus, the prevalent {\em correlational} approach does not allow {\em causal} conclusions free of attribute confounds and often does not include all intersectional groups that might be of interest. An exception is a recent experimental method to evaluate intersectional stereotypes in language models by systematically varying language \cite{charlesworth2024extracting}. 

We introduce an experimental method to assess intersectional stereotypes in VLMs by systematically varying both language {\em and} images, taking advantage of the recent development of generative methods in AI. Since we focus on the social perception of face images, we employ a recently published synthetic dataset of face images that were generated using a generative adversarial network (GAN)~\citep{karras2017progressive}, where a number of protected attributes (race, gender, age) and additional visual attributes (pose, facial expression, and lighting) are manipulated systematically and independently of each other~\citep{liang_benchmarking_2023}. We call this dataset {\em CausalFace}. Using CausalFace, we carried out the experimental evaluation of social biases in CLIP, a popular large vision-and-language model, and we compared our findings with those that may be obtained using the prevalent observational technique based on face images that are sampled in the wild and annotated.

While previous studies of bias in CLIP mostly rely on observational data, a few have explored the use of synthetic images.
\citet{smith2023balancing} use generative models to change people's gender in wild-collected images to de-bias their dataset and address the issue of spurious correlations. This study does not directly control confounding attributes and, therefore, does not allow for intersectional analysis. \citet{wolfe_evidence_2022} generate morphed face images to examine mixed-race associations in CLIP. In our study, we maintain clear distinctions between races without exploring racial mixing.
Concurrent studies \cite{fraser2024examining, howard2024social} also utilize counterfactual, synthetic images to investigate biases related to gender and race in the image perception of VLMs. Our approach differs in several ways, most notably by focusing specifically on face-cropped images and emphasizing the intersectionality of gender, race, age, and facial expression.

\section{Methods}
\subsection{Measuring social perception}
\label{sec:measuring-bias}

We assess social perception in VLMs by computing the cosine similarities between text embeddings and image embeddings, following CLIP's image retrieval metric~\citep{radford_learning_2021}.  
Using this metric, we calculate the similarity between a set of image embeddings $\mI$ (e.g., all face images of a demographic group) and a set of words $D$ that represent a social perception dimension (see \autoref{tab:attributes}) as 
\begin{equation}
\label{eq:cossim}
\cos(\mI, D) = \mean_{\vi \in \mI, d \in D, t \in T} \left( \cos(\vi, \vt(d, t)) \right),
\end{equation}
where $\vt(d, t)$ is the text embedding of a prompt obtained by splicing a social trait adjective $d$ into a prompt template $t$  out of a predefined set of prompts $T$ (e.g., ``a photo of a $<$adjective$>$ person'', see \autoref{tab:templates} for all templates). Averaging similarities over a set of templates is a common practice to make results more robust~\citep{berg_prompt_2022}.

In addition, we introduce a new metric: the difference in cosine similarity compared to a neutral text prompt, i.e., when no adjective (denoted as ``$\varnothing$'') is inserted in the template:
\begin{align}
\label{eq:cossim-delta}
&\Delta \cos(\vi,d) = \mean_{t \in T} \left( \cos(\vi, \vt(d, t)) - \cos(\vi, \vt(\varnothing, t) \right) \notag \\
&\Delta \cos(\mI, D) = \mean_{\vi \in \mI, d \in D} \left( \Delta \cos(\vi,d) \right).
\end{align}

We introduced this metric because we observed a bias between demographic groups when using neutral text prompts. When we consider the distributions of cosine similarities of CausalFace images using a neutral text prompt (e.g., ``a photo of a person''), we find that Asian women have significantly higher and Black men significantly lower cosine similarities compared to all other groups (p $<=$ 0.01). Subtracting neutral prompt cosine similarities normalizes away this ``ground'' bias and more clearly reveals social biases 
(see \autoref{appendix:neutral} for more details).

All results were retrieved using CLIP ViT-B/32 as provided by OpenAI through the \texttt{openai-clip} Python library.

\autoref{tab:templates} shows the full set of contextualized prompt templates used in this study. The dimensions sourced from theories of social perception used in this study are shown in \autoref{tab:attributes}.

\begin{table}[t!]
\centering
\small
\caption{Prompt templates used to infer text embeddings.}
\label{tab:templates}
\begin{tabular}{l}
\toprule
A photo of a $<$adjective$>$ person.\\
A $<$adjective$>$ person.\\
This is a $<$adjective$>$ person.\\
Cropped face photo of a $<$adjective$>$ person.\\
\bottomrule
\end{tabular}
\end{table}
\begin{table*}[ht!]
\centering
\small
\caption{\textbf{Adjectives per Dimension.} \textmd{The Stereotype Content Model (SCM) and the ABC model specify a set of psychometrically validated adjectives for each dimension.}}
\label{tab:attributes}
\resizebox{\textwidth}{!}{%
\begin{tabular}{cc|cccccc}
\toprule
\multicolumn{2}{c}{SCM}   & \multicolumn{6}{c}{ABC}                                                                  \\
\cmidrule(l{2pt}r{2pt}){1-2} \cmidrule(l{2pt}r{2pt}){3-8}
Warmth      & \multicolumn{1}{c}{Competence}  & Pos. Agency    & Neg. Agency   & Progressive Belief         & Conservative Belief     & Pos. Communion & Neg. Communion  \\
\midrule
warm        & competent   & powerful    & powerless  & science-oriented & religious    & trustworthy & untrustworthy \\
trustworthy & intelligent & high-status & low-status & alternative      & conventional & sincere     & dishonest     \\
friendly    & skilled     & dominating  & dominated  & liberal          & conservative & friendly    & unfriendly    \\
honest      & efficient   & wealthy     & poor       & modern           & traditional  & benevolent  & threatening   \\
likeable    & assertive   & confident   & meek       &                  &              & likable     & unpleasant    \\
sincere     & confident   & competitive & passive    &                  &              & altruistic  & egoistic \\
\bottomrule
\end{tabular}
} 
\end{table*}

\subsection{Image datasets}
CausalFace\footnote{The term ``CausalFace'' was introduced by us, as the authors do not name their dataset.} is a synthetic face dataset introduced by \citet{liang_benchmarking_2023}. Using Generative Adversarial Networks (GANs), the authors create realistic-looking synthetic faces and vary each along six attributes (gender, age, race, lighting, viewpoint, facial expression) to obtain a balanced and controlled test set. 

In order to obtain demographic diversity while balancing all confounding variables, the authors proceeded in three steps. First, 100 random seeds are used to sample as many ``seed faces'' from the GAN's latent space. 
 
Second, six ``prototypes'' are generated from each seed face by varying the seed as proposed by~\citet{balakrishnan2021towards} to obtain each one of three different races (Asian, Black, and White) and binarized genders (women and men), thus producing neutral-expression frontally viewed faces of six distinct synthetic people, each one of which belongs perceptually to a different demographic group. The six prototype faces belong to the same pseudo-identity (or seed) and, thus, are as similar as possible in all other attributes (e.g., facial proportions, clothing, background) except their race and gender. 

Third, each prototype is varied in steps along age (+9 additional variations), smiling (+9), lighting (+7), and pose (+4). 
Consequently, CausalFace contains 30 (1~initial plus 29 variations) face images per prototype. Those variations are performed for each of the six gender-race demographic groups. Therefore, we obtain 180 face images per seed\footnote{This number is larger than the 120 facial images per seed that \citet{liang_benchmarking_2023} used for their analysis on bias in face recognition.}. Overall, we used images from 100 different seeds, six prototypes per seed, and 30 variations per prototype. Consequently, our study uses 18,000 unique images. Additional details, including example images of the data set, can be found in Appendix~\ref{sec:data-viz} 

The important property of this dataset is that face attributes are manipulated systematically and independently, enabling an experimental approach and supporting causal interpretation of the results. To achieve this, the authors visually inspected whether manipulating one attribute induces unwanted changes in other attributes. 
Additionally, for each image, the authors collected nine human annotations per attribute to ensure that the apparent ID of the face was not changing and to check that the manipulations of the various attributes were successful. \citep[][Section 3.5 and Fig 4.]{liang_benchmarking_2023}.  
The averages of the nine rater responses suggest that the manipulations are equally effective across demographic groups, i.e., no manipulation bias was found in CausalFace \citep[][Fig. 3]{liang_benchmarking_2023}.
The annotators were recruited worldwide through Amazon Mechanical Turk, and thus, the base of annotators was diverse.

Generating synthetic datasets that probe finer aspects of facial appearance, e.g. the differences between Mediterranean and Scandinavian physiognomies within the White group, is conceptually possible and may yield additional insights in both CLIP and human social judgments. In generating synthetic faces, one must, of course, pay attention to potential correlates between the variables of interest. For instance, pose and image color are known to confound gender bias~\cite{meister2023gender}. The authors of CausalFace control for unwanted correlates through human annotation of their synthetic faces. They directly address pose and do not investigate image color. Our own examination reveals small differences in brightness between gender-race prototypes (Appendix~\ref{appendix:brightness}). 
Future research is needed to test for this potential confounder and, if necessary, correct any biases.

We used two popular real-world wild-sampled image datasets to compare our findings on CausalFace with findings from conventional observational studies. 

FairFace \citep{karkkainen_fairface_2021} is a real-world image dataset commonly used in studies on AI bias. It includes 108,501 face images labeled by race, gender, and age. It offers a balanced representation across seven racial categories and two genders. Of those categories, we use only the overlapping racial categories of FairFace with the other two datasets, which are Asian, White, and Black. We also included only images that are annotated with an age of 20 years and older. Therefore, we have a subset of 38,744 images from FairFace.

Similarly, UTKFace \citep{zhifei2017cvpr} images are annotated with demographic details such as race, gender, and age. From its 23,708 face images, we used a subset of 14,630 images applying the same criteria as for FairFace. 
Additional details on image selection for both datasets can be found in \autoref{sec:data-viz}.

\subsection{Comparing induced variation}
\label{sec:method-variation}

Which attributes of the face image affect social judgment the most? CausalFace varies each attribute systematically and independently. Thus, it uniquely enables us to measure how differences in legally protected attributes (e.g., age, race, gender), as well as unprotected attributes (facial expressions, lighting, and pose), affect differences in social judgment perception. 
Given the varying number of {\em values} within each {\em attribute} (for instance, ``smiling'' has ten values (\autoref{fig:apx_smile}),
whereas ``gender'' has two values---men and women), we use a sampling strategy to ensure a fair comparison.

Specifically, we randomly choose pairs of images with two distinct values, $x_1,x_2 \subset X$, of the chosen attribute $X$ (e.g., images with values White and Black for attribute race) and obtain the respective CLIP image embeddings, $\vi_1(x=x_1), \vi_2(x=x_2)$.
Importantly, all other attributes except the varied one are held constant (in this example, attribute race is varied, and attributes gender, age, smiling, lighting, and pose are held constant).

We then compute the absolute difference in $\Delta$ cosine similarities of the two image embeddings w.r.t. a social dimension $D$, defined as $\AbsDiff(\vi_1, \vi_2, D) = \lvert \Delta \cos(\vi_1, D) - \Delta \cos(\vi_2, D) \rvert$.
This process is repeated 1,000 times for each of the eight social dimensions, generating a bootstrap distribution of 8,000 $\AbsDiff$ values for each of the seven attributes. 
These distributions describe the impact of the specific attribute on the cosine similarity of image embeddings with the valenced text embeddings.
For ordinal attributes such as age and smiling, we introduce additional constraints. To ensure a perceptually significant change in image appearance, we mandate that two samples must differ by at least a certain threshold. These are set at 1.1 for age and smiling. In other words, we set these constraints so as not to pick two images that look too similar.
The impact of the thresholds can be more easily understood by studying the scale of smiling and age, depicted in 
\autoref{fig:apx_age} and \autoref{fig:apx_smile}.
The method described in this section specifically refers to the results discussed in \autoref{fig:compare_diff}.

\section{Results}

\paragraph{CausalFace images are statistically similar to real photographs}
CausalFace images look very realistic. However, they are artificial, and it is reasonable to be concerned that they may differ from photographs of real people in ways that may affect our analysis. Therefore, we explored whether the statistical properties of CausalFace CLIP embeddings differ systematically from those of photographs of real people. To this end, we compared CausalFace to two observational datasets, FairFace~\citep{karkkainen_fairface_2021} and UTKFace~\citep{zhifei2017cvpr}, with respect to six commonly used metrics: mean cosine similarities, markedness \cite{wolfe_markedness_2022}, WEAT \cite{steed_image_2021}, skew@k, max skew@k, and NDKL \cite{geyik_fairness-aware_2019}. 
We find that the CausalFace measure of these six metrics is similar to those in datasets of real images 
(\autoref{tab:bias_metrics}).
This gives us confidence that the CausalFace images are appropriate for our study.
Detailed descriptions of these bias metrics can be found in Appendix~\ref{sec:conventional_metrics}.

\paragraph{Non-protected attributes cause as much social perception variation as protected ones}
\label{sec:results-fedex}

CausalFace allows us to systematically compare variations in perception due to legally protected attributes (e.g., age, race, and gender) as well as unprotected attributes that are usually not included in bias analyses.

\begin{figure}[t]
    \centering
  \includegraphics[width=.8\linewidth]{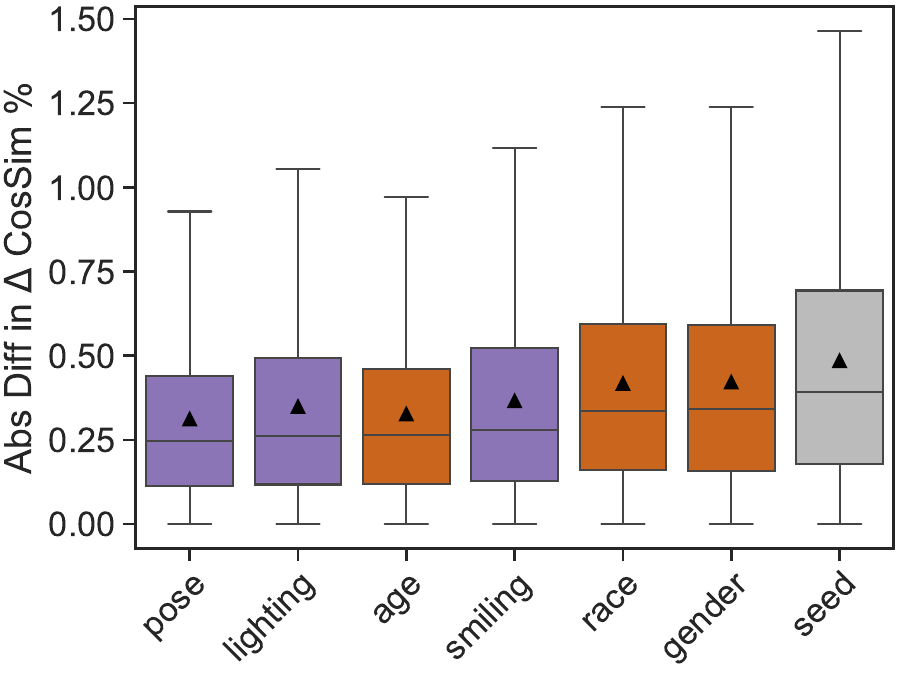}
\caption{\textbf{Protected attributes (orange) and non-protected ones (purple) cause comparable variation in the CLIP embedding.} \textmd{Variation is quantified using a bootstrapping method, calculating the absolute value of $\Delta$ cosine similarity differences between two images compared to each of the valenced text prompts used in our study. The two randomly selected images are distinct for exactly one attribute. For instance, to generate the box plot for ``smiling'', two images of the same pseudo-identity with varying smiling values are compared. The gray box plot (seed) illustrates variation when pseudo-identities are changed while other attributes are maintained constant. 
Boxes are arranged by median values (lines), and means are indicated by triangles.}}
\label{fig:compare_diff}
\end{figure}

\autoref{fig:compare_diff} aggregates the statistics of $\Delta$ cosine similarities between CLIP embeddings of pairs of face images, which are identical apart from variation across the chosen attribute (x-axis). Legally protected attributes are shown in orange, and non-protected attributes are shown in purple.
For comparison, the gray violin represents the variation across different seeds (``pseudo-identities'') within the same demographic groups. The largest variation occurs across different identities (Wilcoxon Rank-Sum test, independent samples, $p<0.001$), meaning faces/identities within a demographic group cause more variation than different demographic groups of the same pseudo-identity.

Furthermore, smiling introduces more variation than age ($p<0.001$), and lighting introduces as much variation as age ($p=0.06$).
In other words, protected and non-protected attributes cause comparable amounts of variation. 
Thus, considering a wide spectrum of protected and non-protected variables is necessary to understand and measure biases comprehensively. CausalFace is unique in providing the necessary attribute annotations.

\paragraph{Facial expression shapes social perception whereas pose and lighting impact legibility}

Previously, we explored the impact of non-protected variables at an aggregate level, averaging across all dimensions of social perception. Next, we shift the focus to a more detailed analysis.
Varying lighting conditions and head orientation do not exhibit a clear trend: Higher $\Delta$ cosine similarities are observed under flat lighting, intermediate values for natural lighting, and lower values for unusual lighting. Similarly, for pose, higher $\Delta$ cosine similarities are found for extreme poses and lower values for central poses. For both variables, negative and positive valences show similar patterns and are positively correlated ($r_{\mathrm{lighting}}=0.93$, $r_{\mathrm{pose}}=0.79$, \autoref{fig:confounds-a}  center and right panel). 

This suggests that changing angles of light incidence (lighting) and head orientations (pose) do not significantly alter social perception but instead affect ``legibility'': the ease with which different social perception dimensions can be interpreted or distinguished. 

Conversely, varying facial expressions show a clear trend: As a face transitions from a frowning expression to a smile, it is perceived as more positive, more Progressive (Belief), and less negative. In other words, the opposing valences are negatively correlated ($r_{\mathrm{smiling}}$=-0.21, 
\autoref{fig:confounds-a} left panel). 

The finding that smiling positively influences social perception has two interpretations:
First, it should be noted that facial expressions are unlabeled in widely used observational datasets like FairFace and UTKFace. Facial expression—and other non-protected variables—may correlate with protected attributes; therefore, ignoring such variables may compromise conclusions from observational studies on algorithmic bias.
Second, it suggests that CLIP exhibits human-like social perception. 
CLIP can classify faces by race and gender 
(\autoref{fig:kde-corrections}), 
demonstrating its ability to make {\em broad} associations related to human perception. The just presented results go one step further and uncover CLIP’s ability to perform {\em fine-grained} social judgments.

\paragraph{Different impact of facial expression on social perception across intersectional groups}

\begin{figure*}[t]
    \includegraphics[width=\textwidth]
    {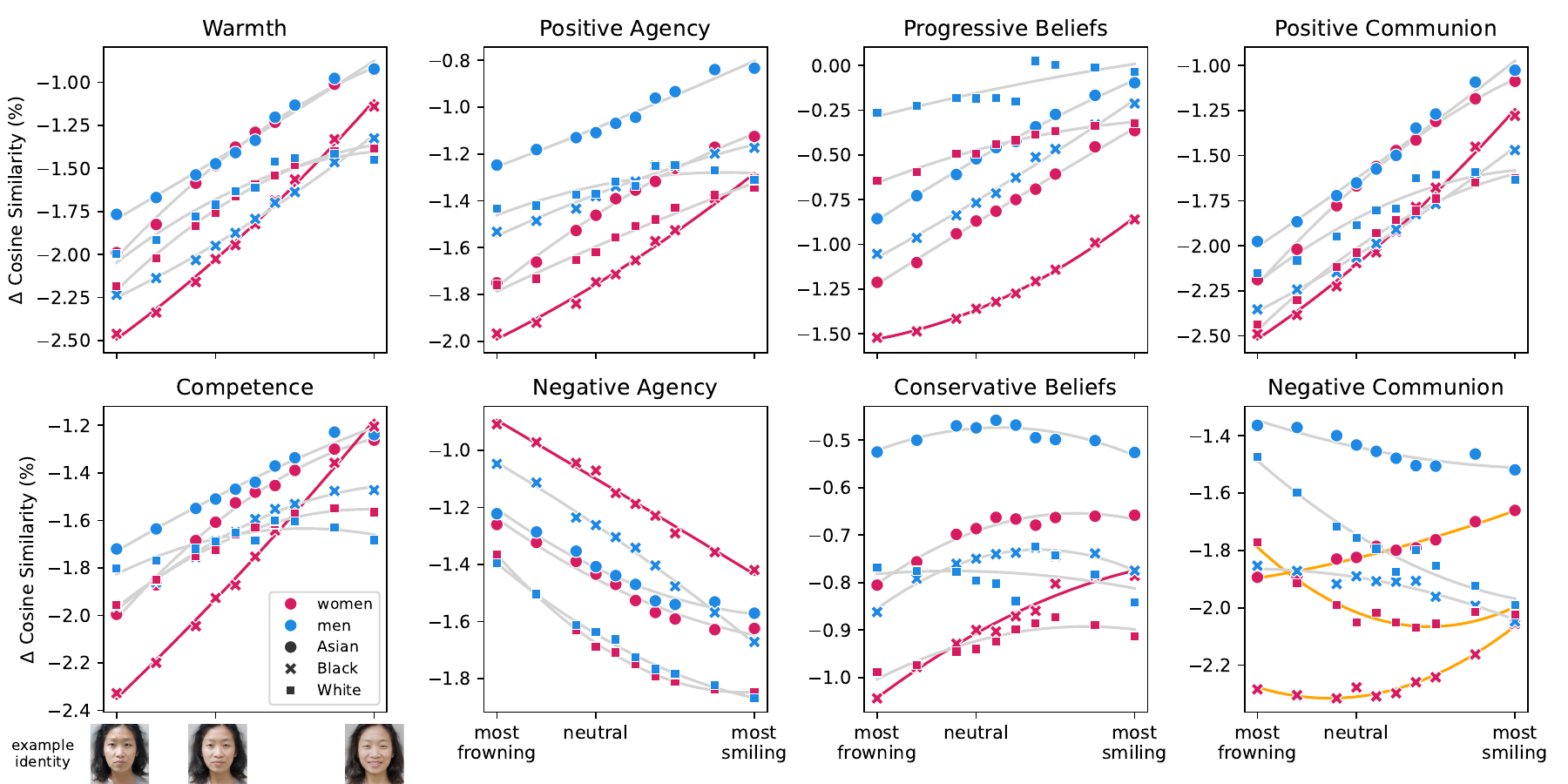}
    \caption{{\bf Facial expression (smiling) influences social perception differently across intersectional groups.} \textmd{Social perception (y-axis) for eight dimensions (panels) plotted against smiling level (x-axis). 
    Social perception is quantified as $\Delta$ cosine similarity, the difference between the cosine similarity of the face image with a relevant valenced-text prompt and the cosine similarity of the image with a neutral prompt (see \autoref{eq:cossim-delta} in the Methods section). Lower $\Delta$ values mean lower similarity between the social perception dimension (specified above each panel) and the corresponding facial expression.
    The means for each set of images sharing the same gender and race are marked with shapes (indicating race) and colors (indicating gender). A second-degree polynomial is fit to the means of each of the six intersectional race-gender sets and is shown in gray. CLIP's perceptions of Black women's faces are marked with red lines to highlight their consistent positioning at the extreme ends in most social perception dimensions. Furthermore, Positive Agency and Negative Communion reveal a clear distinction between men and women. Additionally, in Negative Communion, women (highlighted in orange) show a different trend compared to men.}}
    \label{fig:intersect-smiling}
\end{figure*}

\autoref{fig:intersect-smiling} explores social bias as a function of facial expression.  
It shows that Asian men display the highest values in three social dimensions, namely Positive Agency, Conservative Belief, and Negative Communion.
Another demographic group that stands out even more consistently is Black women (red x lines in \autoref{fig:intersect-smiling}), who consistently appear at the extremes of cosine similarity. 
In all but one panel (Negative Agency), the faces of frowning Black women show the lowest $\Delta$ cosine similarity. However, in four of these panels (Warmth, Positive Communion, Competence, and Conservative Belief), the Black women curve is also the steepest as smiling increases, implying that the metric increases more dramatically than for other gender-race groups. 
As a consequence, most smiling faces of Black women have the highest $\Delta$ cosine similarity for Competence and third-highest for Positive Communion. 
Indeed, the least-smiling Black female faces have the lowest competence cosine similarity of any intersectional group, \emph{and} the most-smiling Black female faces have the highest. Our findings suggest that Black women are an outlier group by CLIP.

Another surprising observation from \autoref{fig:intersect-smiling} is that only one panel---the bottom-right one for Negative Communion---shows different trends for different intersectional groups. 
Specifically, increased smiling leads to decreased Negative Communion {\em only for men}.
In contrast, among Asian and Black women, more smiling paradoxically corresponds to an increase in Negative Communion perceptions. 
For White women, increased smiling first decreases and then increases Negative Communion perception. 
In previously published work investigating human subjects, gender was found to affect honesty judgments in non-smiling individuals but not for smiling individuals \cite{krys2016careful}. 
By contrast, we do not find gender differences in the perception of {\em honesty} (which is similar to ``sincere'', part of the Positive Communion dimension); however, we find it in the perception of {\em dishonesty} (part of Negative Communion). 
Thus, our findings highlight an interesting gap in the literature investigating bias in humans and thus provide a specific hypothesis to be tested further.

\paragraph{CausalFace reveals clear intersectional clusters for race, gender, and age}
\begin{figure}[t!]
    \centering
    \includegraphics[width=\columnwidth]{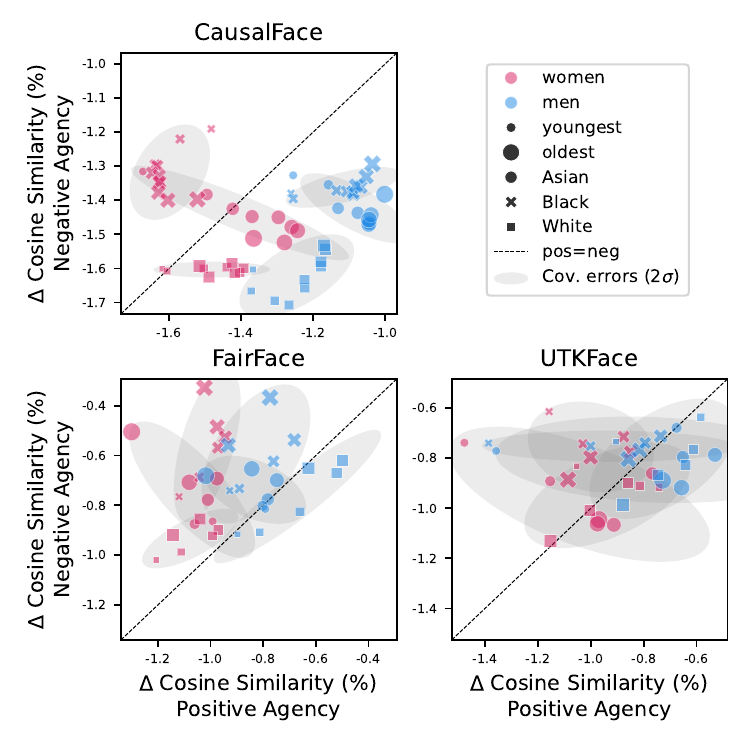}
    \caption{{\bf Social perceptions of intersectional groups are distinct in CausalFace. Not so in FairFace and UTKFace}. \textmd{Markers depict social perceptions separately for gender (marker color), race (marker shape), and age (marker size increasing with age).
    Covariance ellipses (2$\sigma$) for each gender-race intersectional demographic group are shaded in gray. Groups are more sharply defined and distinct in CausalFace, with less overlap in their covariance error ellipses. Negative vs Positive Agency shown here, additional social perception dimensions in \autoref{fig:intersect_age_all}.}} 
    \label{fig:intersect}
\end{figure}

How do the experimental results from CausalFace compare with the traditional bias findings in observational datasets? 
To answer this question, \autoref{fig:intersect} plots $\Delta$ cosine similarities for positive against negative valence attributes of the Agency dimension. Results for each of the ten age values within each of the six intersecting groups are plotted. Covariance error ellipses (2$\sigma$) encompassing each group are shaded in gray. In the top left panel (CausalFace), most of the demographic group ratings are clearly separated in the positive-negative space. Upon closer examination, it is also evident that there is an almost perfect order of the markers within each group from smallest to largest, representing youngest to oldest ages. 

The distinction between groups is less clear in the two wild-collected datasets (bottom panels), as seen visually by the larger sizes and more overlap among different ellipses. The ordering by age within each ellipse is also less evident in those two datasets.

It is likely no coincidence that these differences emerge between the experimentally constructed and wild-collected face images. Most likely, the clusters are separate in the CausalFace representation because facial expression, lighting, and pose are held constant. 
FairFace and UTKFace, on the other hand, do not offer this level of control. 
Our experiments show that non-protected characteristics introduce significant variations (\autoref{fig:compare_diff}). Crucially, these variations may not manifest uniformly across different intersecting identities, underscoring the importance of deploying an experimental approach to measuring bias in VLMs.

\begin{figure}[t!]
    \centering
    \includegraphics[width=\linewidth]{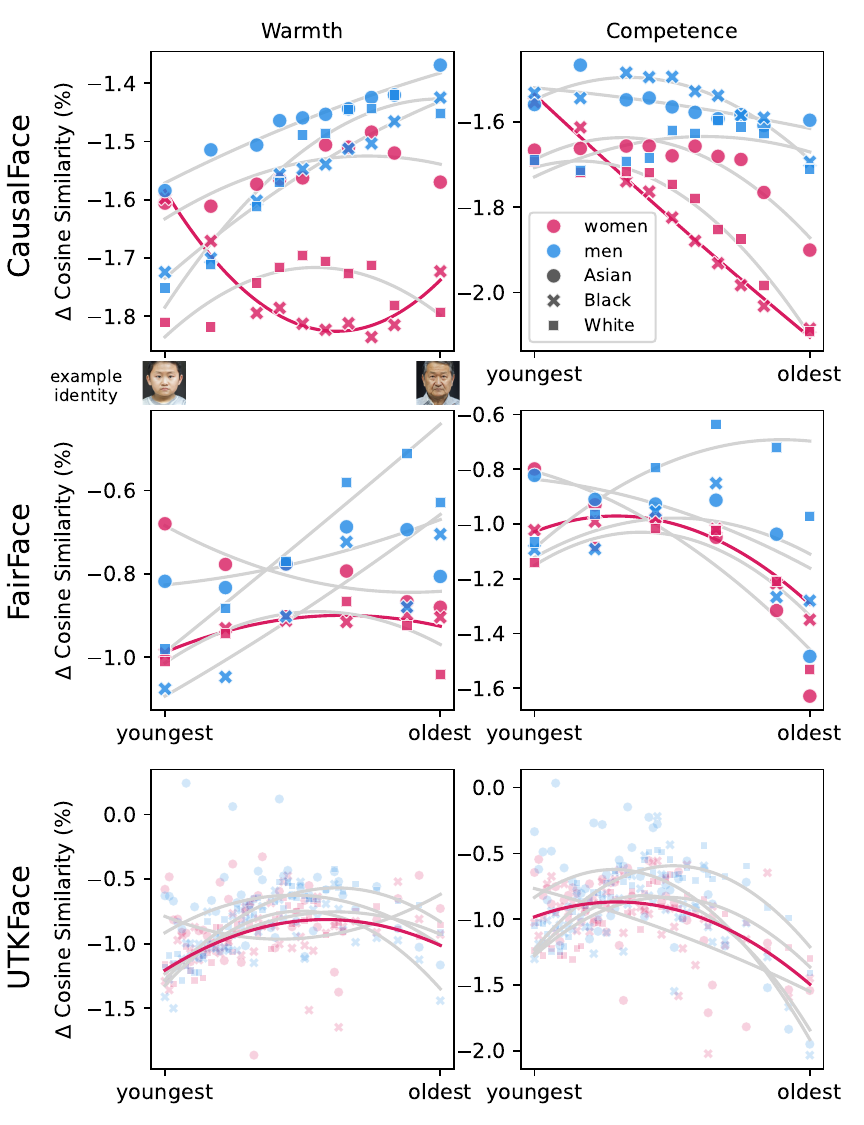}
    \caption{CausalFace shows different trait trajectories vis-a-vis age as compared to FairFace and UTKFace. 
    \textmd{Markers depict social perceptions of different intersectional groups.
    A second-degree polynomial is fitted for each group. 
    Social perception is quantified as $\Delta$ cosine similarity, highlighting the difference in similarity of a valenced-text prompt from a neutral prompt. When testing images in CausalFace, Black women, highlighted by a red line, display distinct patterns: a U-shape trajectory instead of an inverse U-shape (Warmth) or extreme values (Warmth, Competence). Black women are perceived as most competent when young, which sharply and consistently decreases until old age. 
    These trends are far less clear from FairFace (mid-row) and UTKFace (bottom row).
    Additional social perception dimensions across age are plotted in Figs.~\ref{fig:intersect_age_causalface}, \ref{fig:intersect_age_fairface}, and \ref{fig:intersect_age_utkface}.}}
    \label{fig:intersect_age}
\end{figure}

We now analyze how social perception in CLIP is affected by age: Across all three datasets, we plot age on the x-axis against social perceptions on the y-axis. A second-degree polynomial is fit to perceptions for each race-gender group. \autoref{fig:intersect_age} shows only the results for Agency. 
The other social perceptions are shown in Figs.~\ref{fig:intersect_age_causalface}, \ref{fig:intersect_age_fairface}, \ref{fig:intersect_age_utkface}. 
Unlike facial expressions, age allows us to compare results across all three datasets. 

In the Warmth panel (top-left) of \autoref{fig:intersect_age}, Black women stand out because of a U-shaped pattern in how warmly they are perceived, the opposite of the inverted U-shaped pattern for White women. In other words, Black (White) women are seen as most (least) warm at young ages, least (most) warm in mid-age, and then increasingly (decreasingly) warm in old age.

The distinct U-shaped pattern that Black women show in their Warmth perception extends to Positive Agency and Communion (\autoref{fig:intersect_age_all}).
This consistent U-shaped trajectory across different dimensions might indicate the aspects of the ``strong Black woman'' stereotype as age increases \cite{baker2015reconceptualizing}, although it is not present for competence. Further investigation is crucial to understand and interpret this trend.

The Competence panel (top-right) of \autoref{fig:intersect_age} shows an interesting pattern: CLIP's perceived competence of White and Black women declines with age, most steeply for the latter, while men are roughly constant. This pattern is familiar: in our earlier analysis of facial expressions (\autoref{fig:intersect-smiling}), Black women exhibited a unique trajectory across various degrees of smiling. The clear age decline again shows that Black women are an outlier among intersectional groups, although in this case, white women show a similar pattern.

Some of the specific age-related patterns we find diverge from human subject ratings.
One study reported a decline in perceived Warmth for White women transitioning from young to middle-aged adulthood \cite[p.8]{chatman2022agentic}---we find a similar decrease for Black women but not for White women. The same study found no change in Warmth perception for aging White men, whereas we observed an increase for older men across all three races \cite[p.8]{chatman2022agentic}. 

Unlike facial expressions, age allows us to compare results across all three datasets. Intriguingly, the age trends for Black women we see in CausalFace are not evident in FairFace nor in UTKFace. This suggests that uncontrolled attributes in FairFace and UTKFace make for noisy measurements and may hide interesting phenomena.

\section{Discussion}
We investigated whether the popular CLIP vision-language model is able to make social judgments from photographs of human faces and whether attributes of such images affect social perception. We quantified social perception by comparing CLIP embeddings of face images with embeddings of text prompts from two theories of social psychology. Our test images come from a synthetic dataset---CausalFace---in which image attributes were varied systematically and independently, allowing us to investigate the effect of each attribute on social perception, as well as the relative importance of legally protected and non-protected image attributes that are not labeled in observational datasets. 

We find that CLIP exhibits human-like social perception. Similarly to human observers, it will classify faces according to race and gender, and increased smiling leads to more positive social perception. It is sensitive to age and relatively insensitive to viewpoint and lighting.  
CLIP thus mirrors humans in drawing trait inferences from face images~\cite{willis2006first}. 
Parallels between human and AI perception have been pointed out for Large Language Models (LLMs). They include similarities between word embeddings and human semantic memory~\cite{digutsch2023overlap}, including group stereotypes \cite{rogers2021primer, lewis2020gender, caliskan2017semantics, garg2018word}. 
However, unlike LLMs, which capture meaning from language, an already symbolic description, CLIP works with raw images where meaning is implicit in the pixel brightness values. While previous studies have shown that vision-language models (VLMs) can replicate certain human perceptions, such as susceptibility to visual illusions \cite{zhang2023grounding}, our study is the first to directly demonstrate that these models can also reflect human-like social perceptions and stereotypes.

CLIP's response to age is complex. It exhibits unique age-related patterns that are peculiar to Black women.
The specific age-related patterns for women and men appear different from those found in human studies \cite{chatman2022agentic}.
It would be premature to draw firm conclusions on whether CLIP differs significantly in age perception from humans: human studies use fewer categories than we did and do not control for unlabeled potential confounds, as CausalFace does.

CLIP's social perception of age categories and facial expressions exhibits a unique trend for one gender-race group: Black women. The social perception of Black women becomes more negative from youth until mid-age, instead of becoming more positive as it does for other groups. Additionally, smiling positively affects their social perception more strongly than other groups.
The perception of Black women by human subjects and by AI algorithms has attracted attention in the literature. 
Their gender is recognized more slowly and erroneously by 4-8yo children recruited in the US, identifying as either monoracial White (49\%) or biracial \cite{lei2020bfgender}. 
Similarly, early studies on algorithmic bias reported that commercial classifiers misgendered darker-skinned women more frequently than other demographics~\cite{buolamwini2018gender} (a later study found those observations to be better explained by bias in the test sets~\cite{balakrishnan2021towards}).
Another study suggests that semantic differential categorization of Black women does not appear to add up race and gender components, as is true for other race-gender pairs~\citep{billups2022intersectionality}. 
Moreover, when describing events, speakers tend to mention people who are more like them first. This ``like me'' effect occurs across various populations but does not apply to Black women \cite{brough2024cognitive}. This linguistic uniqueness is hypothesized to stem from a combination of historical stereotypes, prejudices, pressures, and resistances \cite{higginbotham1992african}.

We draw two lessons that will be useful to other researchers. First, it is important to control all main facial attributes when studying face perception. We find that facial expression impacts social perception more than age, and lighting has as much impact as age. This implies that studies of bias vis-a-vis protected attributes that do not control for other (unprotected) attributes may reach the wrong conclusions. Indeed, when we repeat our analysis using two popular public datasets of wild-collected face images, where facial expression, lighting, and pose are not controlled, demographic biases are no longer apparent. These observations, made possible by our experimental method, are novel and have not been previously reported.

Second, generative AI offers the opportunity to produce good-quality synthetic stimuli and thus will help researchers move from observational to experimental approaches. An experimental approach offers considerable advantages for testing both AI systems and human subjects: controlled, systematic, incremental manipulation of variables across a large number of gender-race-age combinations allowed us to discover the new phenomena just described. Additionally, our conclusions are causal, thanks to mutually independent control of the variables of interest. For instance, varying facial expressions of Black women elicit the most extreme changes in social perceptions in CLIP as compared to other gender-race intersections. This bias pattern had not been previously observed, likely because variations from wild-collected datasets are perturbed by noise from uncontrolled confound variables, and thus, some effects are hidden.

Our study is relevant to social psychology because it demonstrates that VLMs can generate testable hypotheses, thereby enabling more efficient piloting and targeted resource allocation for human studies.
A body of growing literature shows that LLMs replicate human behavior \cite{strachan2024testing,mei2024turing} and thus are increasingly used to replace humans in behavioral studies \cite{strachan2024testing,agnew2024illusion} and/or generate testable hypotheses~\cite{strachan2024testing,mei2024turing,bail2024can}. Our study shows that this is also viable for VLMs.

CausalFace does not account for all intersections of protected and non-protected attributes. For instance, different age levels are available only for a single type of facial expression, lighting, and pose. Generating and studying a richer set of attribute intersections may reveal further interesting phenomena. Finally, our findings refer to one specific CLIP variant and should, therefore, not be naïvely generalized to all VLMs or to all AI systems. We hope, nevertheless, that our method will provide a useful blueprint to study the behavior of current and future VLMs that are trained using different datasets, model architectures, and training procedures.

\section*{Acknowledgements}
Carina I. Hausladen and Colin F. Camerer were supported by NSF DRMS grant 1851745.
Manuel Knott was supported by an ETH Zurich Doc.Mobility Fellowship. Pietro Perona received support from the Allen E. Puckett Chair at Caltech.

We extend our gratitude to Ralph Adolphs, Umit Keles, Matthew Monfort, Robert Wolfe, Angelina Wang, Eliza Stenzhorn, Neehar Kondapaneni, and Laure Delisle for their helpful comments and discussions. 
Additionally, we appreciate comments provided by participants of various conferences where we presented the paper. We also thank our colleagues for their in-depth thoughts during internal lab meetings at the Camerer Lab at Caltech, and the Chair for Computational Social Science at ETH Zurich.

{
\small

\bibliographystyle{unsrtnat}
\bibliography{ref_bibfish}
}

\clearpage
\begin{appendices}
\onecolumn
\setcounter{figure}{0}   
\setcounter{table}{0} 
\renewcommand{\thesection}{\Alph{section}}%
\renewcommand{\thesubsection}{\thesection.\arabic{subsection}}
\renewcommand\thefigure{S.\arabic{figure}} 
\renewcommand\thetable{S.\arabic{table}}
\begin{center}
    \begin{minipage}{\textwidth}
        \centering
        \fontsize{16}{20}\selectfont
        Supplementary Materials
    \end{minipage}
\end{center}
\vspace{.5cm}

\section{Conventional bias metrics}
\label{sec:conventional_metrics}

Since CausalFace contains artificially generated images, one could question the dataset's representativeness, as these faces might have different properties with respect to bias compared to real photos.
Therefore, in the following, we compare CausalFace to the two observational datasets concerning six metrics: WEAT score, markedness, Skew@k, MaxSkew@k, and NDKL.

\subsection{Adapted SC-WEAT}

The traditional WEAT (word embedding association test) \citep{caliskan2017semantics} requires opposite pairs of target items. Although these pairs have been validated in human linguistics, their antonymic relationship in CLIP's embedding space remains unconfirmed. Given this uncertainty, we opted for the SC-WEAT, which obviates the need to define such pairs.
In the following, we draw upon \citet{steed_image_2021}, who adapt the WEAT for investigating vision-language cosine similarities. More specifically, we adapt the Single-Category WEAT (SC-WEAT) as follows:
Let $\mA$ and $\mB$ be two sets of image embeddings of equal size (e.g., men and women), and let $\vd \in \mD$ be a single text embedding from a certain social perception dimension $\mD$ (e.g., Warmth). 
Then, we can calculate the test statistic for single text embedding as follows:

\begin{equation}
    s(\vd, \mA, \mB) = \mean_{\va \in \mA}\cos(\vd, \va) - \mean_{\vb \in \mB}\cos(\vd, \vb)
\end{equation}

\begin{equation}
    s(\mD, \mA, \mB) = \mean_{\vd \in \mD} s(\vd, \mA, \mB)
\end{equation}

The test statistic measures the differential association of the target $\vd$ concept with the image attributes $\mA$ and $\mB$. The effect size (measured in Cohen's $d$) is:

\begin{equation}
    es(\vd, \mA, \mB) = \dfrac{s(\vd, \mA, \mB)}{\std_{\vz \in \mA \cup \mB} \cos(\vd, \vz)}
\end{equation}

\begin{equation}
    es(\mD, \mA, \mB) = \mean_{\vd \in \mD}es(\vd, \mA, \mB)
\end{equation}

We test the significance of this association with a permutation test over all possible equal-size partitions $\{ (\mA_i , \mB_i ) \}_i$ of $\mA \cup \mB$ to generate a null hypothesis as if no biased associations existed. The one-sided $p$-value measures the likelihood of obtaining the test result at least as extreme as the observed one, under the assumption that the null hypothesis (there is no bias) is true:

\begin{equation}
p = \Pr[s(\mD, \mA_i, \mB_i) > s(\mD, \mA, \mB)]
\end{equation}

\autoref{tab:bias_metrics} shows the WEAT scores for CausalFace and the two observational datasets. 
The most striking similarities in means are observed when comparing Asians to Blacks and Whites to Blacks. 
In the comparison of Male versus Female, CausalFace and FairFace demonstrate closely matched scores, while UTKFace records a lower WEAT value. 

\subsection{Markedness}

The concept of {\em markedness} was thoroughly reviewed by \cite{wolfe_markedness_2022}.
The metric is defined as the percentage preference for a neutral prompt over an attribute-specific prompt.
Consider an image featuring a person with a specific attribute, such as being ``white.'' Our results show that the image category ``white'' has high markedness. This means that when an image of a white face is embedded in CLIP, the model prefers the descriptor ``a photo of a person'' over ``a photo of a white person.''
In contrast, low-markedness categories include Black and Asian. For these image categories, CLIP prefers a specific prompt (``a photo of a Black person'') over a neutral prompt (``a photo of a person'').
Intuitively, these results show that ``white'' is the default racial category in CLIP.

\begin{itemize}
    \item Let $\vi$ be the embedding of an image of a person with the given attribute.
    \item Let $\vt_n$ be the embedding of an unmarked/neutral text prompt, e.g., ``a photo of a person.''
    \item Let $\vt_m$ be the embedding of a marked text prompt, e.g., ``a photo of a \textit{white} person.''
    \item Calculate the cosine similarity: $\cos(\vi, \vt_n)$ and  $\cos(\vi, \vt_m)$.
    \item Repeat this process for all images in the dataset categorized under the given attribute (e.g., \textit{white}).
    \item Count the number of times in which $\cos(\vi, \vt_n) > \cos(\vi, \vt_m)$. Markedness describes the fraction of images where this condition is true.
    \item With $N$ being the total number of images of persons with the given attribute in the dataset, we can express the Markedness percentage $M\%$ as:

\begin{equation}
    M\% = \frac{1}{N} \sum_{i=1}^{N} 1_{\{\cos(\vi, \vt_n) > \cos(\vi, \vt_m)\}} \times 100
\end{equation}
\end{itemize}

In all three datasets, CLIP shows a preference for an unspecified prompt over one that specifies the race as ``White.'' The percentages are 51.15\%,  54.12\%, and 32.58\%. CausalFace falls in the middle, exhibiting neither the highest nor the lowest preference for an unspecified term. CausalFace and FairFace both display a significant decrease in preferring an unmarked prompt for the remaining four race or gender categories. 
Compared to UTKFace, this decline is more pronounced in FairFace than in CausalFace (\autoref{tab:bias_metrics}). 

\subsection{Mean cosine similarities} 
Mean cosine similarities are calculated as the mean of cosine similarities between an image category and all positive items of the social perception dimensions (\autoref{eq:cossim}).
The mean cosine similarities exhibit uniform trends across all three datasets. Specifically, Asians consistently display higher cosine similarities than Whites, who exhibit higher similarities than Blacks (\autoref{tab:bias_metrics}).

\subsection{Skew@k, MaxSkew@k, NDKL}
We present three additional bias metrics, following \citet{geyik_fairness-aware_2019}.

{\em Skew@k} measures the difference between the desired proportion of image attributes in a ranked list $\tau$ and the actual proportion \cite{geyik_fairness-aware_2019}. Specifically, $\tau_k^T$ represents the top $k$ items in the ranked list $\tau$ for a given text query $T$.
For example, given the text query ``this is a friendly person,'' a desired proportion of the image attribute gender could be 50\%. 
Let the desired proportion of images with gender/race label $A$ in the ranked list be $p_{d,T,A} \in [0, 1]$, and the actual proportion be $p_{\tau_T, T, A} \in [0, 1]$. 
The resulting Skew of $\tau_T$ for an attribute label $A \in \mathcal{A}$ is 
\begin{equation}
    Skew_A@k(\tau_T) = \ln \frac{p_{\tau_T, T, A}}{p_{d, T, A}}
\end{equation}
In other words, Skew@k is the (logarithmic) ratio of the proportion of images having the gender/race value $a_i$ among the top $k$ ranked results to the corresponding desired proportion for $A$. A negative Skew@k corresponds to a lesser than desired representation of images with gender/race $A$ in the top $k$ results, while a positive Skew@k corresponds to favoring such images.

\textit{MaxSkew@k} describes the maximum skew among all attribute values \citep{geyik_fairness-aware_2019}.

The {\em NDKL} (Normalized Discounted Cumulative KL-Divergence) employs a ranking bias measure based on the Kullback-Leibler divergence, measuring how much one distribution differs from another \citep{kullback1951information}. 
This measure is non-negative, with larger values indicating a greater divergence between the desired and actual distributions of attribute labels for a given text query $T$. 
It equals 0 in the ideal case of the two distributions being identical for each position. Let $D_{\tau_i}^T$ and $D^T$ denote the discrete distribution of image attributes race/gender in $\tau_i^T$ and the desired distribution, respectively. NDKL is defined by 
\begin{equation}
    NDKL(\tau_T) = \frac{1}{Z} \sum_{i=1}^{|\tau_y|} \frac{d_{KL}(D_{\tau_{i_T}} \parallel D_T)}{\log_2(i + 1)}
\end{equation}
where 
\begin{equation}
    d_{KL}(D_1 \parallel D_2) = \sum_j D_1(j) \ln \frac{D_1(j)}{D_2(j)}
\end{equation}
is the KL-divergence of distribution $D_1$ with respect to distribution $D_2$, and $Z = \sum_{i=1}^{|\tau^r|} \frac{1}{\log_2(i+1)}$ is a normalization factor. The KL-divergence of the top-$k$ distribution and the desired distribution is a weighted average of Skew@k measurements (averaging over $A \in \mathcal{A}$).

The leftmost block of the bottom subtable in \autoref{tab:bias_metrics} depicts {\em Skew@k}. Asians predominantly receive positive values, indicative of a favorable social perception relative to other races. The values for Blacks, however, are distinctly negative.
In the case of Whites, CausalFace consistently shows negative social perception, while FairFace and UTKFace mostly hover around zero.
In contrast, Skew@k for men and women hovers near zero, implying no significant bias towards either gender.

{\em MaxSkew@k} offers a slightly different perspective. Regarding race and gender, CausalFace aligns more closely with FairFace, with UTKFace appearing as the outlier in this block (with a larger gender value and smaller race value). 

Lastly, {\em NDKL}, akin to MaxSkew, reveals a more pronounced bias for race than for gender. The mean values across the datasets are closely aligned for gender but not for race. Unlike MaxSkew, the mean values for all three datasets appear similarly distant with respect to the NDKL.

Overall, Skew@k, MaxSkew@k, and NDKL indicate that no dataset consistently emerges as an outlier. CausalFace behaves differently only in one metric and for one race from the observational datasets. Therefore, we conclude that the three metrics reinforce the findings discussed in the previous paragraph: CausalFace does not significantly diverge from the observational datasets.

\begin{table*}[ht]
\centering
\caption{\textbf{CausalFace, FairFace, and UTKFace have similar overall statistics in CLIP latent space.} 
{\em Markedness} indicates the preference (in \%) for a neutral prompt over a race or gender-specific one. White is mostly unmarked, while all other categories are marked. 
The {\em mean cosine similarities} (in \%) represent averages of cosine similarities from both the SCM and the positive dimensions of the ABC models. 
{\em WEAT} scores represent the mean difference across all dimensions (e.g., A+).
A negative {\em Skew@k} corresponds to a lesser than desired representation of candidates with gender/race  $A$ in the top $k$ results, while a positive Skew@k corresponds to favoring such images. We set k=1000. 
{\em NDKL} is a non-negative metric and equals 0 in the ideal case.
}
\vspace{5mm}
\label{tab:bias_metrics}

\begin{minipage}{.59\linewidth}
\begin{adjustbox}{max width=\textwidth}
\centering
\begin{tabular}{lcccccc}
\toprule
\raisebox{-1.0ex}[0pt][0pt]{Image} & \multicolumn{3}{c}{Markedness} & \multicolumn{3}{c}{Mean Cosine Similarity} \\
\cmidrule(lr){2-4} \cmidrule(lr){5-7}
Category & CausalFace & FairFace & UTKFace & CausalFace & FairFace & UTKFace \\
\midrule
White & 51.15 & 54.12 & 32.58 & 23.55 & 23.07 & 22.25 \\
Black & 1.00 & 3.92 & 2.89 & 23.22 & 22.61 & 21.93 \\
Asian & 0.05 & 3.88 & 4.07 & 24.07 & 23.42 & 22.68 \\
Male & 2.90 & 0.00 & 20.84 & 23.48 & 23.07 & 22.03 \\
Female & 6.63 & 0.00 & 11.59 & 23.74 & 23.17 & 22.45 \\
\bottomrule
\end{tabular}
\end{adjustbox}
\end{minipage}%
\hfill
\begin{minipage}{.39\linewidth}
\centering
\begin{adjustbox}{max width=\textwidth}
\begin{tabular}{lrrr}
\toprule
\raisebox{-1.0ex}[0pt][0pt]{Image} & \multicolumn{3}{c}{SC-WEAT} \\
\cmidrule(lr){2-4} 
Category & CausalFace & FairFace & UTKFace \\
\midrule
White-Black & 0.31 & 0.34 & 0.24 \\
Asian-Black & 0.68 & 0.64 & 0.60 \\
Asian-White & 0.43 & 0.29 & 0.37 \\
Male-Female & -0.11 & -0.08 & -0.36 \\
\bottomrule
\end{tabular}
\end{adjustbox}
\vspace{2pt}
\end{minipage}
\begin{adjustbox}{max width=\textwidth}
\begin{tabular}{l rrr rrr rrr}
    \toprule
    & \multicolumn{3}{c}{Skew} & \multicolumn{3}{c}{MaxSkew} & \multicolumn{3}{c}{NDKL} \\
    \cmidrule(lr){2-4} \cmidrule(lr){5-7} \cmidrule(lr){8-10} 
    Image Category & CausalFace & FairFace & UTKFace & CausalFace & FairFace & UTKFace & CausalFace & FairFace & UTKFace \\
    \midrule
    White   & -0.63 & 0.00 & 0.13 & -    & -    & -    & -    & -    & -    \\
    Black   & -0.98 & -1.44 & -0.53 & -    & -    & -    & -    & -    & -    \\
    Asian   & 0.69 & 0.52 & 0.20 & -    & -    & -    & -    & -    & -    \\
    Male    & -0.16 & -0.13 & -0.35 & -    & -    & -    & -    & -    & -    \\
    Female  & 0.13 & 0.10 & 0.26 & -    & -    & -    & -    & -    & -    \\
    Gender  & - & - & - & 0.13 & 0.12 & 0.26 & 0.01 & 0.00 & 0.02 \\
    Race    & - & - & - & 0.69 & 0.59 & 0.31 & 0.17 & 0.10 & 0.04 \\
    \bottomrule
\end{tabular}
\end{adjustbox}
\end{table*}

\clearpage
\section{Additional results}

\subsection{Facial expression, lighting, and pose confound social perception}
\label{apx:poselighting}

\begin{figure*}[ht!]
    \centering
        \includegraphics[width=\linewidth]{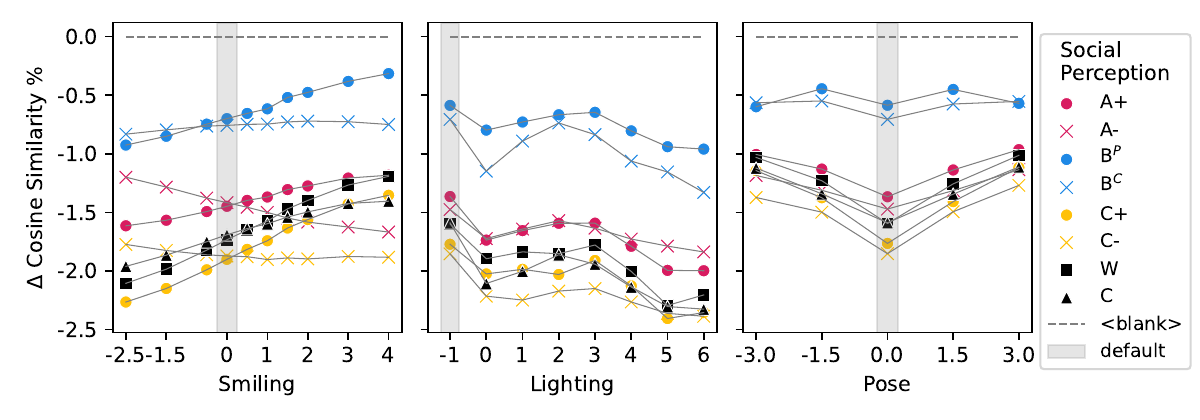}
        \caption{\textbf{Social perception as a function of Smiling, Lighting, and Pose.} Curves for W (Warmth), C (Competence), A+, A- (Agency), B\textsuperscript{P}, B\textsuperscript{C} (Belief), C+, C- (Communion).
        Visualizations of the x-axis values are shown in Figs. \ref{fig:apx_smile}, \ref{fig:apx_pose} and \ref{fig:apx_light}.
        The cosine similarity between an image and a neutral text prompt is used as the reference and is depicted by the horizontal gray dashed line at 0. Markers are obtained by averaging across races and genders. Please see Figs.~\ref{fig:intersect-smiling},\ref{fig:intersect_age} for intersectional plots.
        }
        \label{fig:confounds-a}
\end{figure*}

\autoref{fig:confounds-a} shows how facial expression, lighting, and pose affect social perception dimensions. A gray dashed line marks cosine similarity for a neutral prompt; deviations caused by valence attributes are shown as gray markers. Pearson correlations ($r$) support visual patterns (\autoref{tab:rho-nonprotected}).

Since facial expressions are discussed in the main paper, we focus here on lighting and pose.
The middle panel of \autoref{fig:confounds-a} illustrates the effect of lighting variations on social perception, revealing a consistent pattern across all dimensions. These variations, represented on the x-axis, correspond to different angles of light incidence, visualized in \autoref{fig:apx_light}. 
Although displayed on a linear scale for comparability, lighting variations are not inherently linear, so curve slopes are not central to our analysis.
For all social dimensions, the highest scores are observed when the face is evenly lit (lighting = -1). Other configurations with high scores include lighting from above (lighting = 2 and 3). These findings are consistent with human subject studies \cite{sun1996early, sun1998sun}. In contrast, the lowest scores are observed when the face is lit from below (lighting = 5 and 6), which tends to be perceived as unnatural by humans.

The right panel of \autoref{fig:confounds-a} shows variations of $\Delta$ cosine similarity in pose. We observe a parallel trend in all social perception dimensions. The parallel trend is mirrored by a positive $r$ for all dimensions (average: $r$=0.79).
The curves are mirrored around a pose value of zero, representing a frontal stance. As a face tilts more to the left (pose $<$ 0) or to the right (pose $>$ 0), the curves approach closer to the baseline.
This effect is easier to interpret when considering that the raw (not the $\Delta$) cosine similarities instead. The raw values are the highest for frontal poses and decrease the more an image is tilted to the left or right (shaped like an upward-facing triangle).
In other words, for side-pose photos, the image embedding is more distant from the ``person'' mode in the embedding space. This makes the interpretation of changes induced by adding social adjectives to the text prompt challenging.
This observation suggests that uncontrolled pose variations in the observational, wild-collected datasets are a significant source of noise that makes analyzing other effects difficult and could explain a significant share of the noise present in wild-collected datasets (\autoref{sec:data-viz}). Future research could address this observation in more detail.
Additionally, psychological literature indicates that head orientation (pose) can influence perceived emotions \cite{zhang2020apparent}. 
This would suggest that we should find a different impact on positive and negative attributes, at least for some of the varied head positions. However, in \citet{zhang2020apparent}, the postures affecting emotions were upward and downward head postures, while in CausalFace, we have left-right orientation variation, which may explain the difference in our observations. Additionally, we observe that negative valence is consistently ranked lower than positive valence, unaffected by head position. This suggests that another factor, currently unmeasured and uncontrolled, may affect this aspect.

Overall, this suggests that changing angles of light incidence (lighting) and head orientations (pose) do not significantly alter social perception but instead affect ``legibility'': the ease with which different social perception dimensions can be interpreted or distinguished.

\begin{table}[ht!]
    \centering
    \caption{Pearson correlation $r$ to assess the impact of confounds on social perception. We compare opposite dimension pairs from the ABC model.}
    \label{tab:rho-nonprotected}
    \begin{tabular}{l rr rr rr}
    \toprule
    & \multicolumn{2}{c}{Smiling} & \multicolumn{2}{c}{Lighting} & \multicolumn{2}{c}{Pose} \\
    \cmidrule(lr){2-3} \cmidrule(lr){4-5} \cmidrule(lr){6-7}
    Dimensions & $r$ & $p$-value & $r$ & $p$-value & $r$ & $p$-value \\
    \midrule
    A+/A-     & -0.99 & $<$0.01 & 0.97 & $<$0.01 & 0.99 & $<$0.01 \\
    B\textsuperscript{P}/B\textsuperscript{C} & 0.90 & $<$0.01 & 0.95 & $<$0.01 & 0.40 & 0.51 \\
    C+/C-     & -0.54 & 0.11 & 0.88 & $<$0.01 & 0.98 & $<$0.01 \\
    average       & -0.21 & & 0.93 & & 0.79 & \\
    \bottomrule
    \end{tabular}
\end{table}

\clearpage
\subsection{Ageing influences social perception across intersectional groups}

\begin{figure}[ht]
    \centering
\includegraphics[width=.8\textwidth]{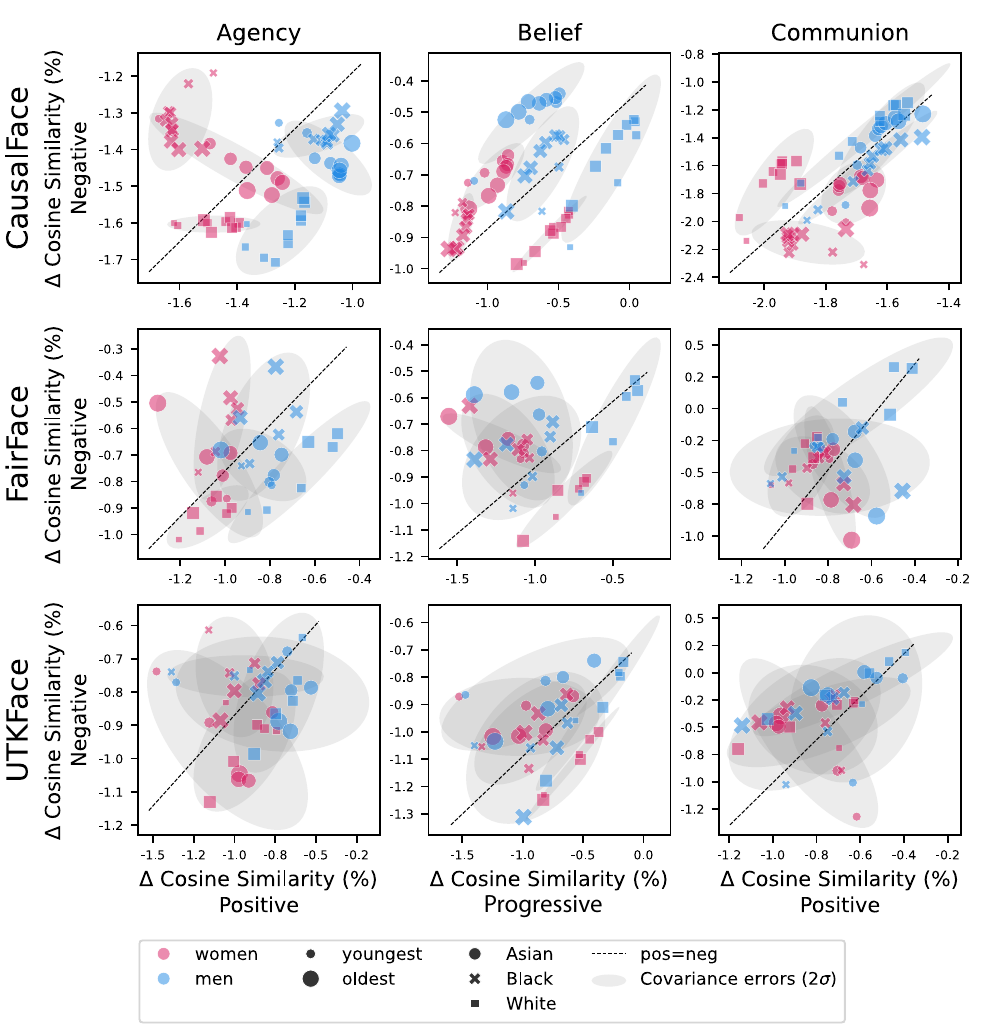}  
    \caption{\textbf{Extension to \autoref{fig:intersect}, additionally displaying the results for Belief and Communion.} Markers depict social perceptions of different intersectional groups. The x-axis represents positive, and the y-axis represents negative valence for Agency, Belief, and Communion. Both axes show the difference in cosine similarities compared to a neutral prompt. Covariance error ellipses (2$\sigma$) for each demographic group (gender + race) are shaded in light gray.}
    \label{fig:intersect_age_all}
\end{figure}

\begin{figure*}[t]
    \centering
    \includegraphics[width=\textwidth]{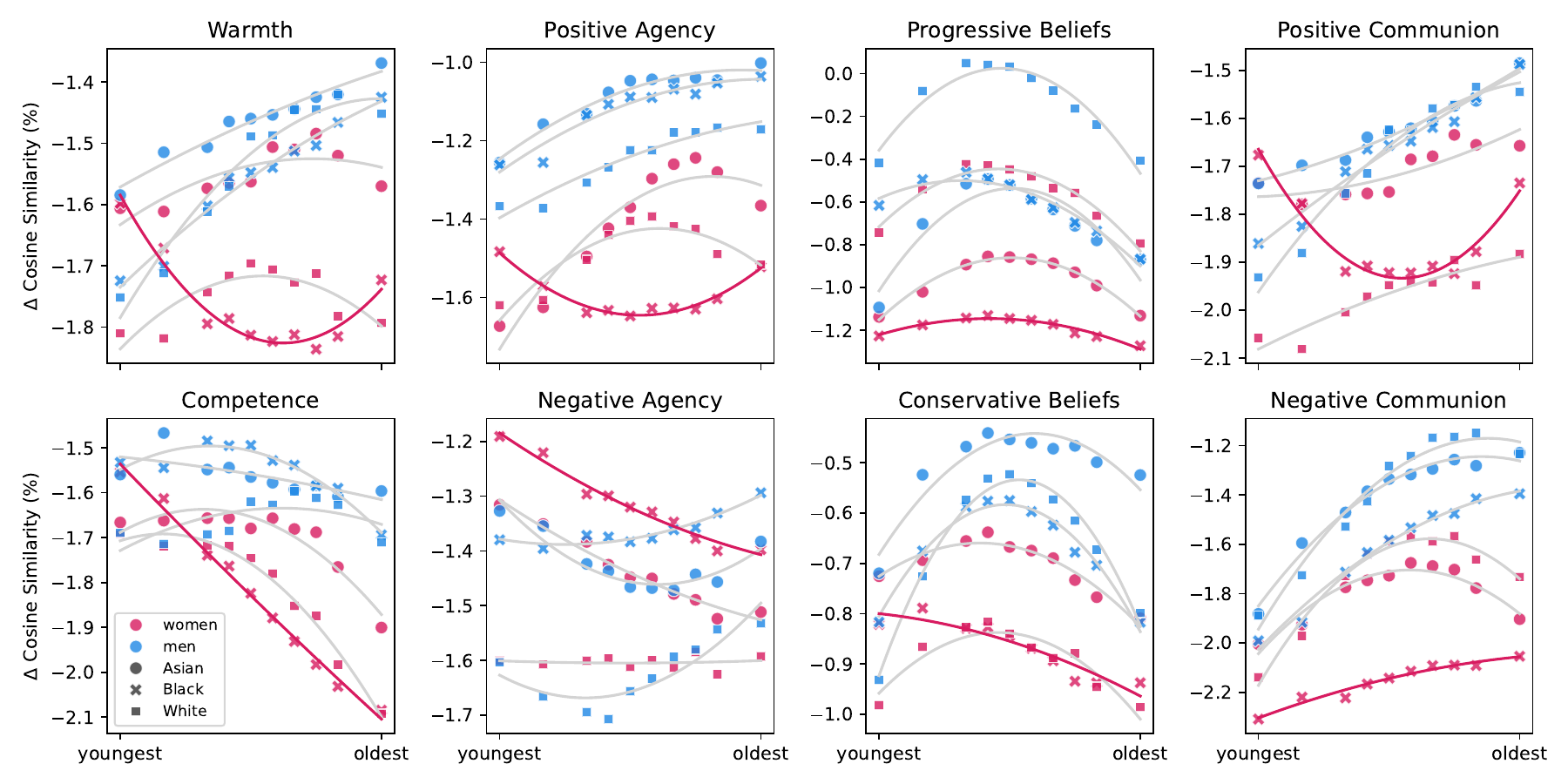}
    \caption{\textbf{Extension to \autoref{fig:intersect_age}, focusing on CausalFace.} Means for each age and race-gender group are marked with shapes (indicating race) and colors (denoting gender). A second-degree polynomial is fitted for each race-gender group. 
    Social perception is quantified as $\Delta$ cosine similarity, highlighting the shift in similarity of an image to a valenced text prompt from a baseline non-valenced prompt.
    CLIP's perception of Black women's faces, marked by a red line, display distinct patterns: a U-shape trajectory instead of the inverse U-shape seen for other demographic groups (Warmth, Positive Agency, Positive Communion),  and extreme values (Competence, Negative Communion, Positive Agency, Progressive Belief). Compare with the same analysis carried out on the in-the-wild observational datasets: FairFace (\autoref{fig:intersect_age_fairface}) and UTKFace (\autoref{fig:intersect_age_utkface}).}
    \label{fig:intersect_age_causalface}
\end{figure*}

\begin{figure}[b]
    \centering
    \includegraphics[width=\textwidth]{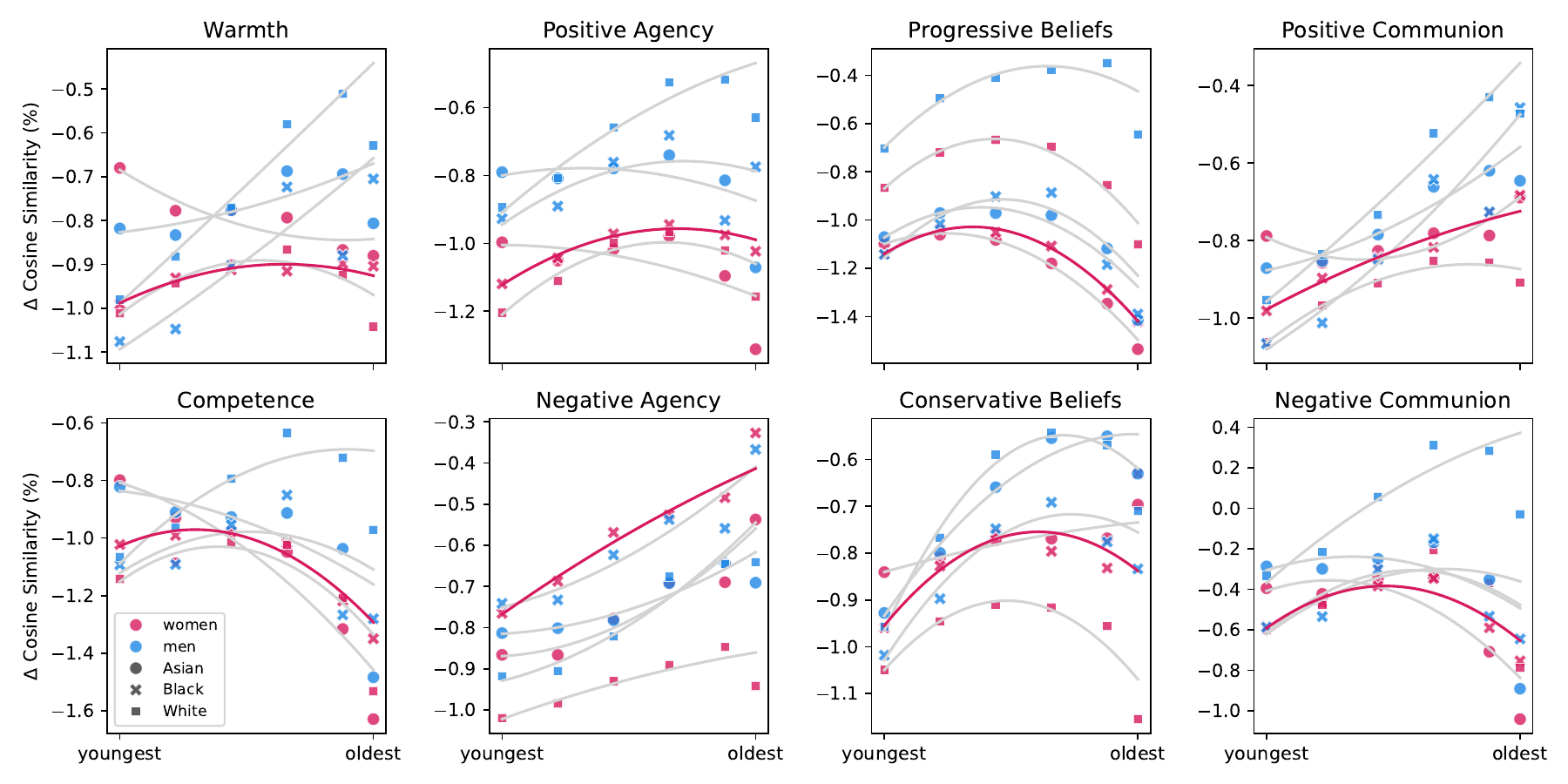}
    \caption{\textbf{Extension to \autoref{fig:intersect_age}, focusing on FairFace.} Data was sampled to ensure an equal number of samples for each age range. Mean values for each age and race-gender group are depicted using shapes (indicating race) and colors (denoting gender). Each group is modeled with a second-degree polynomial. Clusters are identified based on the coefficients from the CausalFace analysis, with each distinguished by unique colors. Black women are highlighted by red lines.}
    \label{fig:intersect_age_fairface}
\end{figure}

\begin{figure}[ht]
    \centering
    \includegraphics[width=\textwidth]{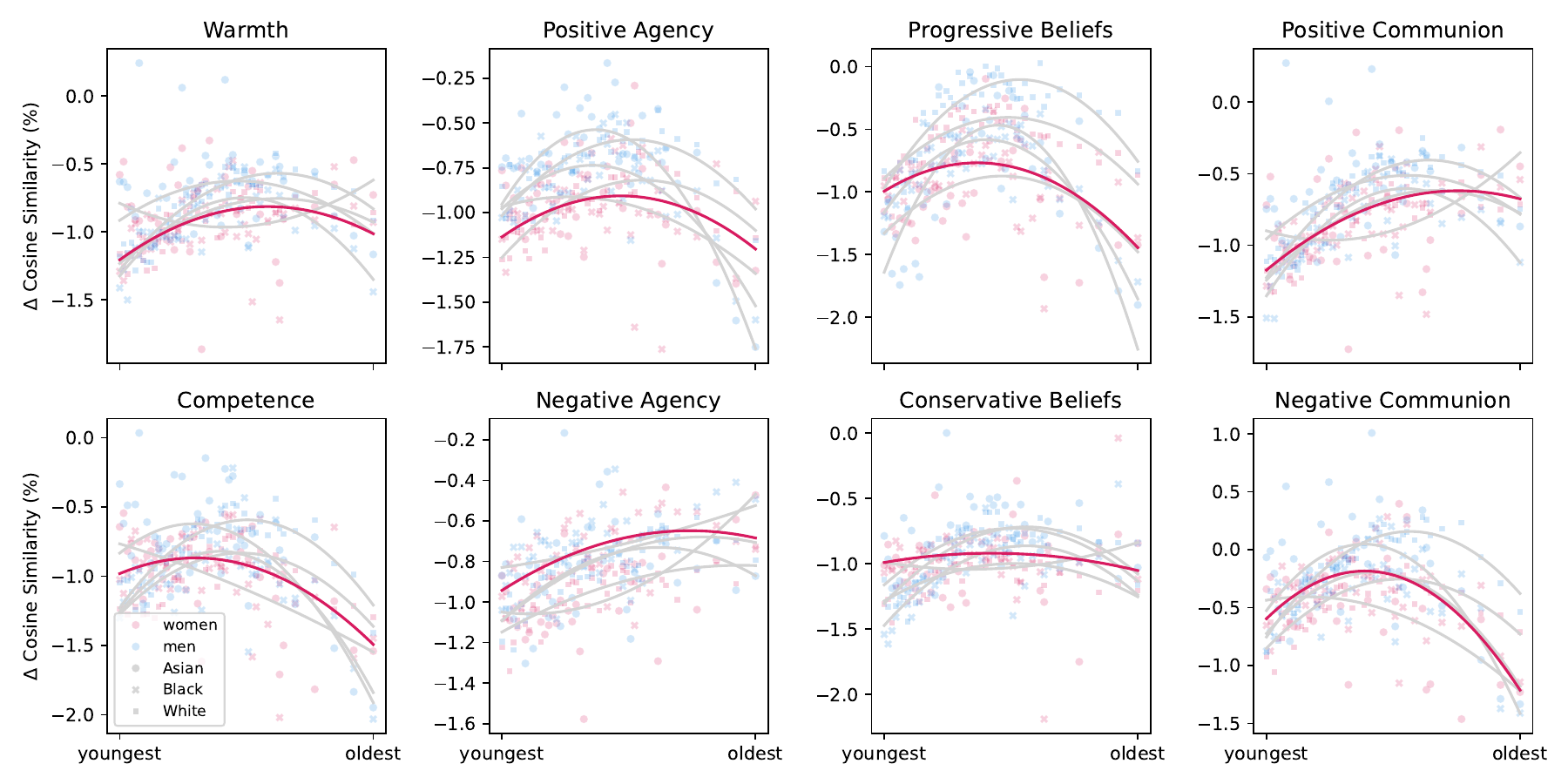}
    \caption{\textbf{Extension to \autoref{fig:intersect_age}, focusing on UTKFace.} Data was sampled to ensure an equal number of samples for each age range. Mean values by age and group are depicted using shapes (for race) and colors (for gender). Each group is modeled with a second-degree polynomial. Clusters are identified based on the coefficients from the CausalFace analysis, with each distinguished by unique colors. Black women are highlighted by red lines.}
    \label{fig:intersect_age_utkface}
\end{figure}

\autoref{fig:intersect_age_all} extends \autoref{fig:intersect} for all social dimensions in the ABC model.
Additionally, we separately plot the positive and negative valence perceptions by age:
Figs.~\ref{fig:intersect_age_causalface},
\ref{fig:intersect_age_fairface} and
\ref{fig:intersect_age_utkface} extend \autoref{fig:intersect_age}.

\FloatBarrier

\clearpage
\section{Dataset details}
\label{sec:data-viz}

\paragraph{FairFace and UTKFace}
Out of the races present in FairFace, we used the original White and Black categories and aggregated East Asian and Southeast Asian into a single category. Indian, Hispanic, and Middle Eastern were not used in this study.
Similarly, for UTKFace, we use Asian, White, and Black, but not the Indian and Other categories.
For both datasets, we only used images annotated with age 20 or older.
This is motivated by matching the general demographic distribution available in CausalFace. 

\paragraph{CausalFace}
The original CausalFace images have a resolution of 512x512 pixels. To emphasize the faces and minimize the influence of the background and clothing, as well as to match the style of UTKFace images, we cropped each image to 432x432 pixels around the faces. This cropping was done by selecting pixels from 40 to 472 horizontally and from 0 to 432 vertically. An exception occurs with the side-view poses (\autoref{fig:apx_pose}), where we horizontally cropped the images from 0 to 432 pixels for negative pose values and from 80 to 512 pixels for positive pose values, to ensure that faces are centered in the image.

\begin{figure}[ht!]
    \centering
    \includegraphics[width=.9\textwidth]{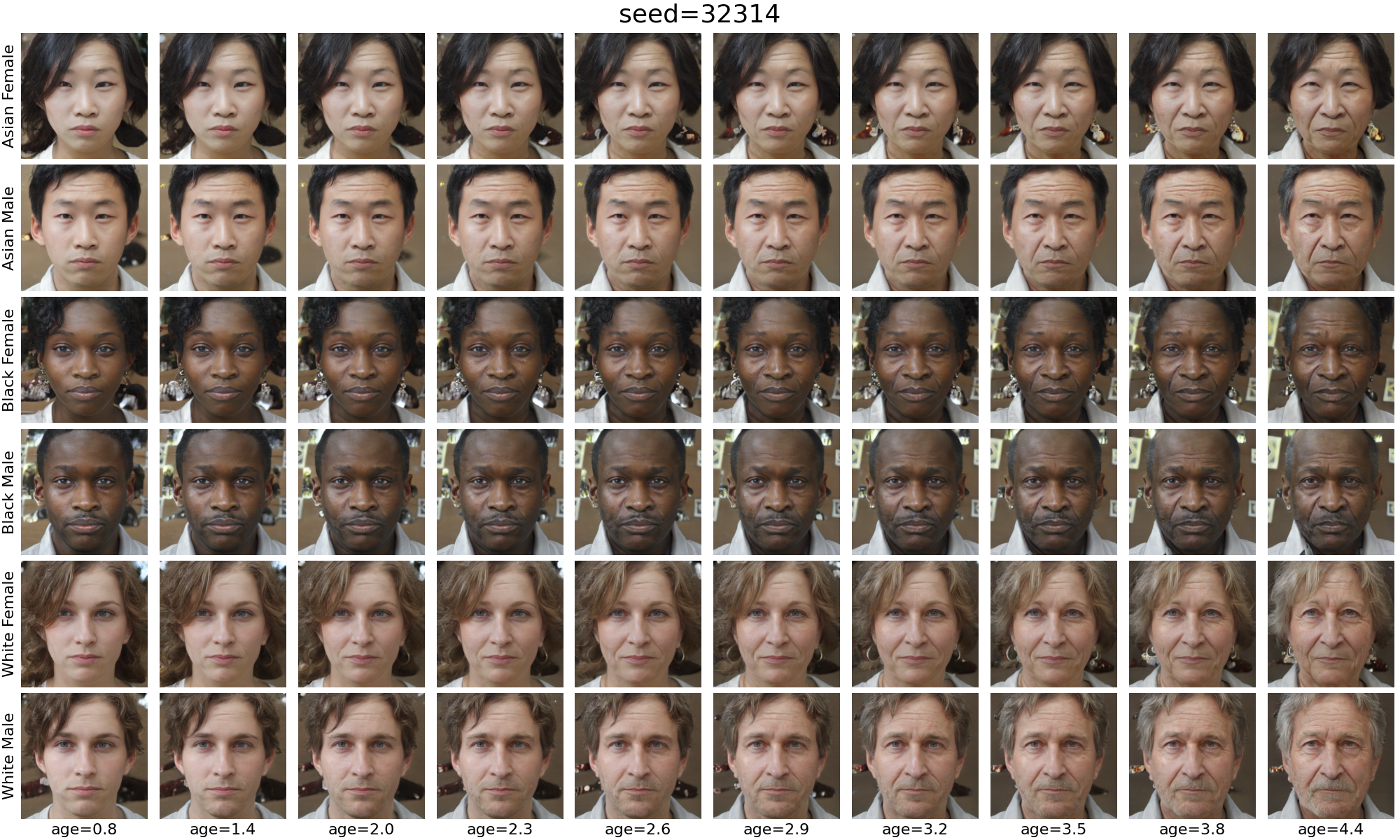}
    \caption{CausalFace example images for a single seed altered by ``age''.}
    \label{fig:apx_age}
\end{figure}

\begin{figure}[ht!]
    \centering
    \includegraphics[width=.9\textwidth]{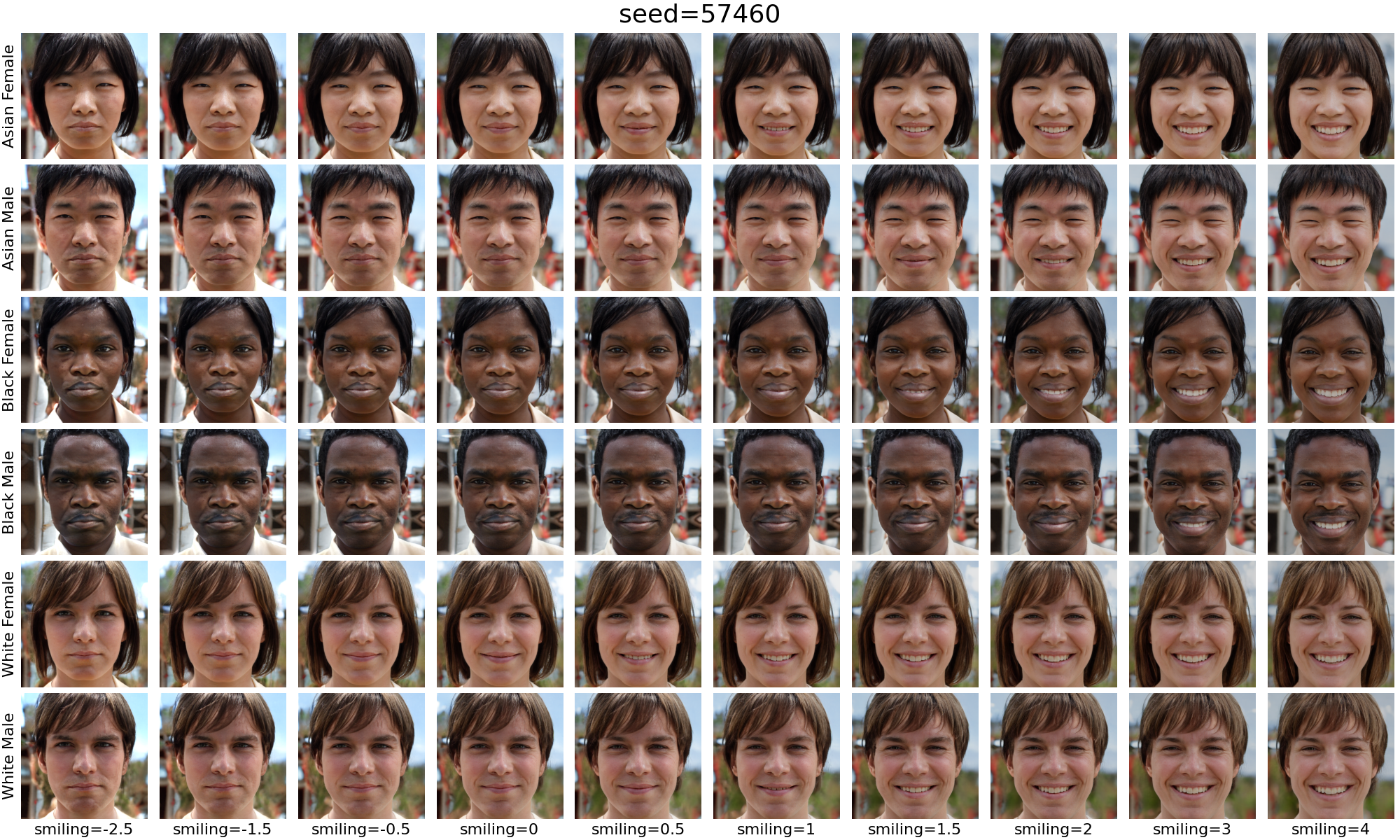}
    \caption{CausalFace example images for a single seed altered by ``smiling''.}
    \label{fig:apx_smile}
\end{figure}

\begin{figure}[ht!]
    \centering
    \includegraphics[width=.5\textwidth]{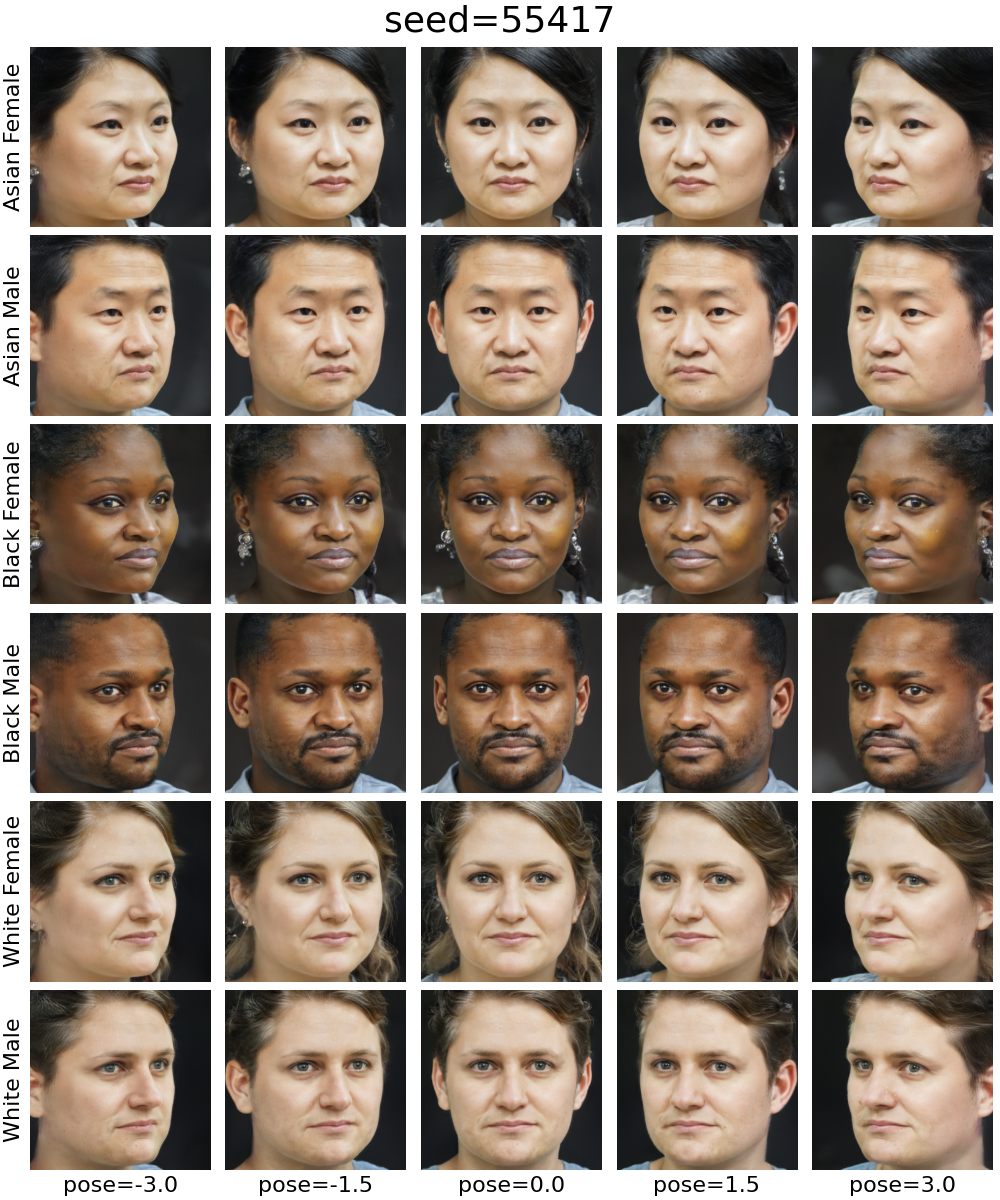}
    \caption{CausalFace example images for a single seed altered by ``pose''.}
    \label{fig:apx_pose}
\end{figure}

\begin{figure}[ht!]
    \centering
    \includegraphics[width=.8\textwidth]{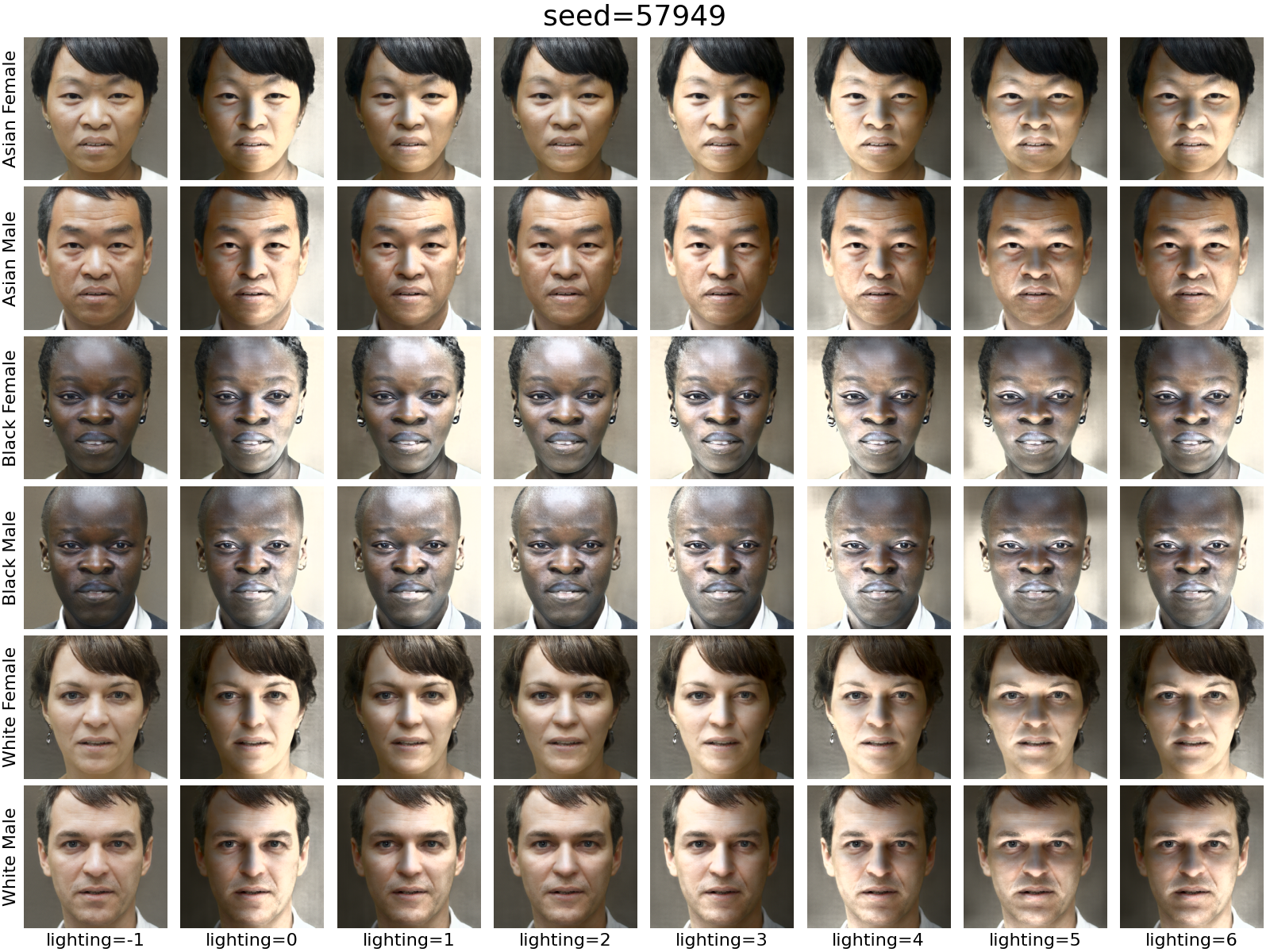}
    \caption{CausalFace example images for a single seed altered by ``lighting''.}
    \label{fig:apx_light}
\end{figure}

We also identified a slight spurious correlation between facial expression (``Smiling'') and the average image brightness in the original CausalFace images. To address this, we made small adjustments to the brightness of images in the Smiling category:
First, we estimate the average brightness in the images' face regions (binary masks for faces are available in CausalFace). Then, we scale the pixel values of ``Smiling'' variants to match the average brightness of the corresponding prototype image (where smiling=0, see \autoref{fig:apx_smile}). We tested our results using both the adjusted and unadjusted images and observed only minor differences in CLIP cosine similarity when inferring with text prompts.
Figs.~\ref{fig:apx_age}-\ref{fig:apx_light} show example images of the CausalFace images used in this study (after adjustments).

\clearpage
\section{Brightness analysis in CausalFace}
\label{appendix:brightness}

\begin{figure}[ht!]
    \centering
    \includegraphics[width=.75\textwidth]{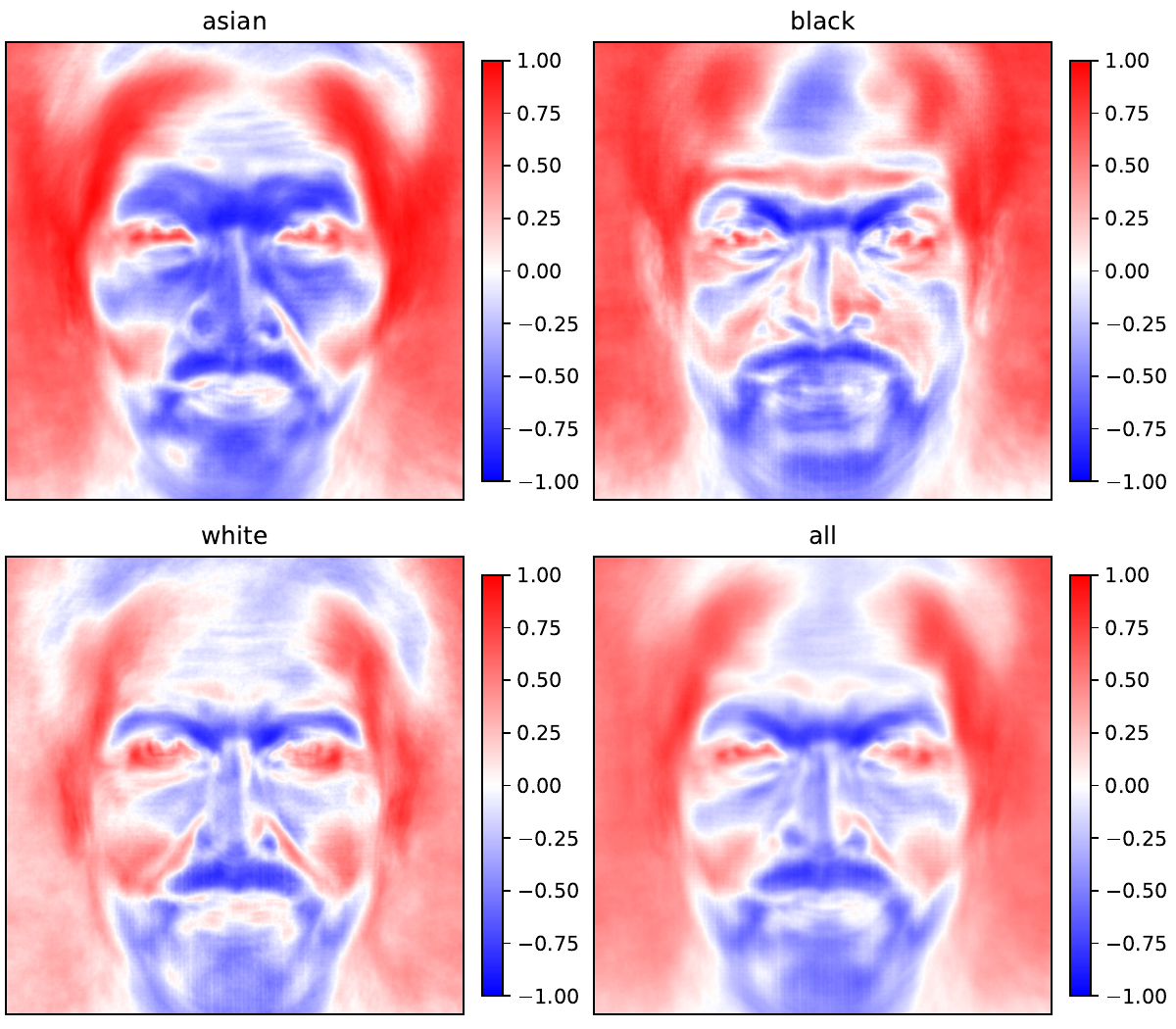}
    \caption{\textbf{Pixel-wise differences in grayscale values between the male and female prototype within a seed aggregated across all seeds (see \autoref{eq:heatmap}).} Red means that these areas are more frequently darker for women, and blue areas are more often darker for men. As expected, the strongest effects are visible in the hair and beard regions. We attribute these regions to gender-specific attributes rather than unwanted confounds.
    We can also see a smaller effect for the background (slightly darker for women) and clothing (slightly darker for men). Even though the differences are small ($\pm4\%$ in brightness), future research could address eliminating these correlations.}
    \label{fig:gender-heatmap}
\end{figure}

\begin{figure}[ht!]
    \centering
    \includegraphics[width=\textwidth]{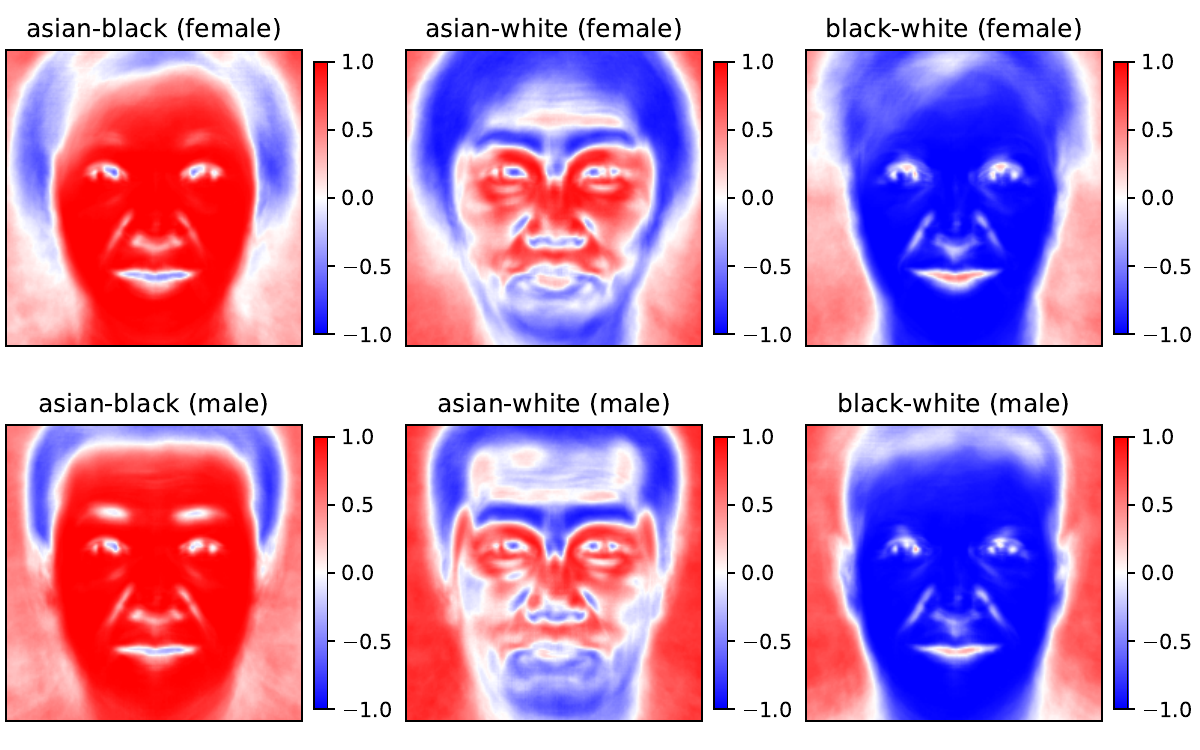}
    \caption{
    \textbf{Pixel-wise differences in grayscale values between two different racial prototypes within a seed aggregated across all seeds (see \autoref{eq:heatmap}).}
    Blue means that these areas are more frequently darker for the first-mentioned race, and red areas are more often darker for the second-mentioned. Besides the expected race-specific attributes (skin and hair color), we can also observe that Asian photos have the lightest background and White photos have the darkest background. In future research, improved synthetic datasets should address these correlations.
   }
    \label{fig:race-heatmap}
\end{figure}

Motivated by the findings of \citet{meister2023gender}, we explore the pixel-wise difference in grayscale values (brightness) between different demographic groups.
More specifically, we examine whether the brightness of a pixel between two images (e.g., the male and female prototype within a seed) is greater, smaller, or equal. We do this for all seeds in the dataset and estimate the frequencies of how often these three cases occur.
Formally, for each pixel position $P$, the displayed value $X_P$ that is shown in Figs.~\ref{fig:gender-heatmap} and \ref{fig:race-heatmap} is calculated as 
\begin{equation}\label{eq:heatmap}
    X_P = \dfrac{1}{n} \sum_{i=0}^{n} x_{P,i}
\end{equation}
\begin{equation}
    x_{P,i} = \begin{cases}
  \phantom{-}1  & Y_{P,i,a} > Y_{P,i,b} \\
  \phantom{-}0  & Y_{P,i,a} = Y_{P,i,b} \\
  -1  & Y_{P,i,a} <  Y_{P,i,b} \\
\end{cases}
\end{equation}

where $n$ is the number of seeds, and $Y_{P}$ is the grayscale value of an image in a given pixel position.
If $X_P = 1.0$ or $X_P = -1.0$, that pixel is darker or brighter for one of the two groups 100\% of the time.  If $X_P = 0.0$, brighter and darker values occur equally often in that position.

\clearpage
\section{Neutral text prompt associations in CausalFace}
\label{appendix:neutral}

\begin{figure}[p]
\vspace{-1cm}
    \centering
    \includegraphics[height=0.9\textheight]{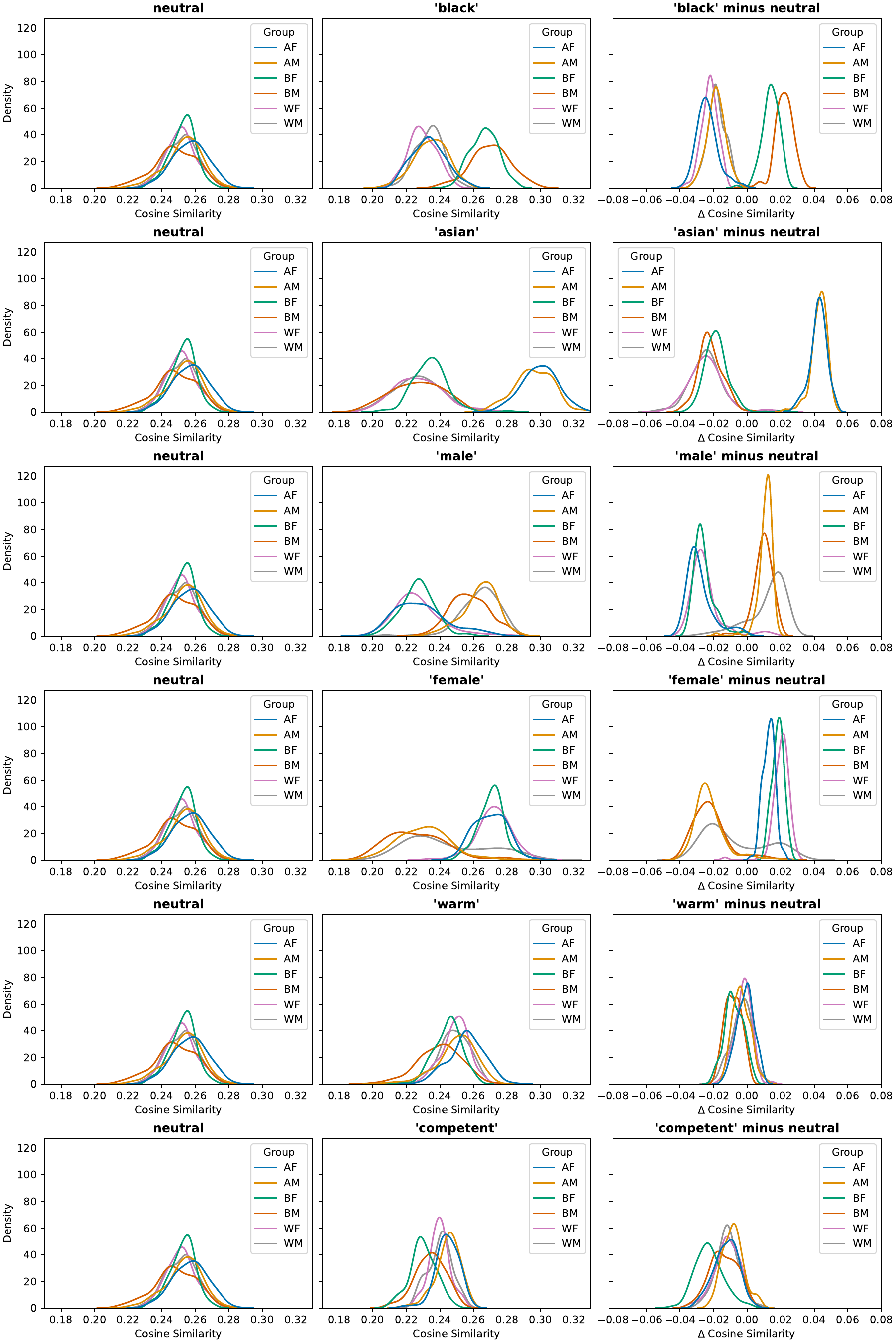}
\caption{ \textbf{The neutral prompt bias.} Probability densities for cosine similarities between text prompts and face images. Cosine similarities are shown separately for six demographic groups (colored lines). N=100 CausalFace face image prototypes (or pseudo-identities) are used for each demographic group (see also \autoref{sec:data-viz}). The probability densities are approximated using kernel densities.
{\em Left column}: Cosine similarities to a neutral prompt (e.g., ``a photo of a person''). Notice that cosine similarities vary across prototype images (see also \autoref{fig:neutral-scatter}). 
{\em Center column}: Cosine similarities to a prompt with a descriptive attribute (e.g., ``a photo of a {\em female} person''). 
{\em Right column}: $\Delta$ cosine similarities after subtracting neutral prompt similarities for each image individually (see \autoref{eq:cossim-delta}), thus removing the bias for the neutral ``person'' association.}
\label{fig:kde-corrections}
\end{figure}

The cosine similarity of face images with neutral prompts (e.g., ``a photo of a person'') is different across images (\autoref{fig:kde-corrections}, left column). To correct this effect, we take the cosine similarity of each face image with the neutral prompt (left column) as the reference point for our evaluations of social perception in CLIP. Thus, to compute the effect of each socially loaded prompt, we subtract the cosine similarity with the neutral prompt from the cosine similarity with the ``socially loaded'' prompt  (\autoref{fig:kde-corrections}, center column) in order to obtain the $\Delta$ cosine similarity (\autoref{fig:kde-corrections} right column). After removing the image-specific neutral prompt bias, the $\Delta$ cosine similarities result in narrower densities and tighter confidence intervals (compare the rightmost with the center column).

\begin{figure}[ht!]
    \centering
    \includegraphics[width=.5\textwidth]{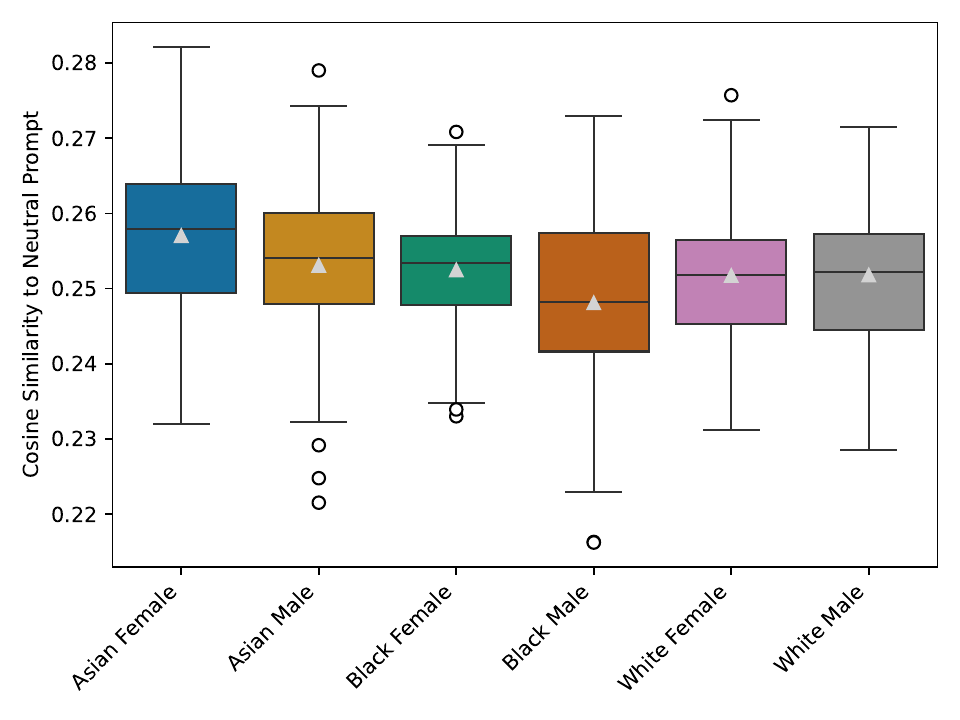}
    \caption{\textbf{Distributions of cosine similarities of CausalFace images (n=100 for each demographic group) with a neutral text prompt (e.g., ``a photo of a person'').} Paired $t$-tests show that Asian women have significantly higher and Black men significantly lower cosine similarities compared to all other groups ($p\leq0.01$).
    We conclude that there exists a bias towards the concept of ``person'' that needs to be accounted for when assessing social biases.}
    
    \label{fig:neutral-boxplot}
\end{figure}

\begin{figure}[ht!]
    \centering
    \includegraphics[width=\textwidth]{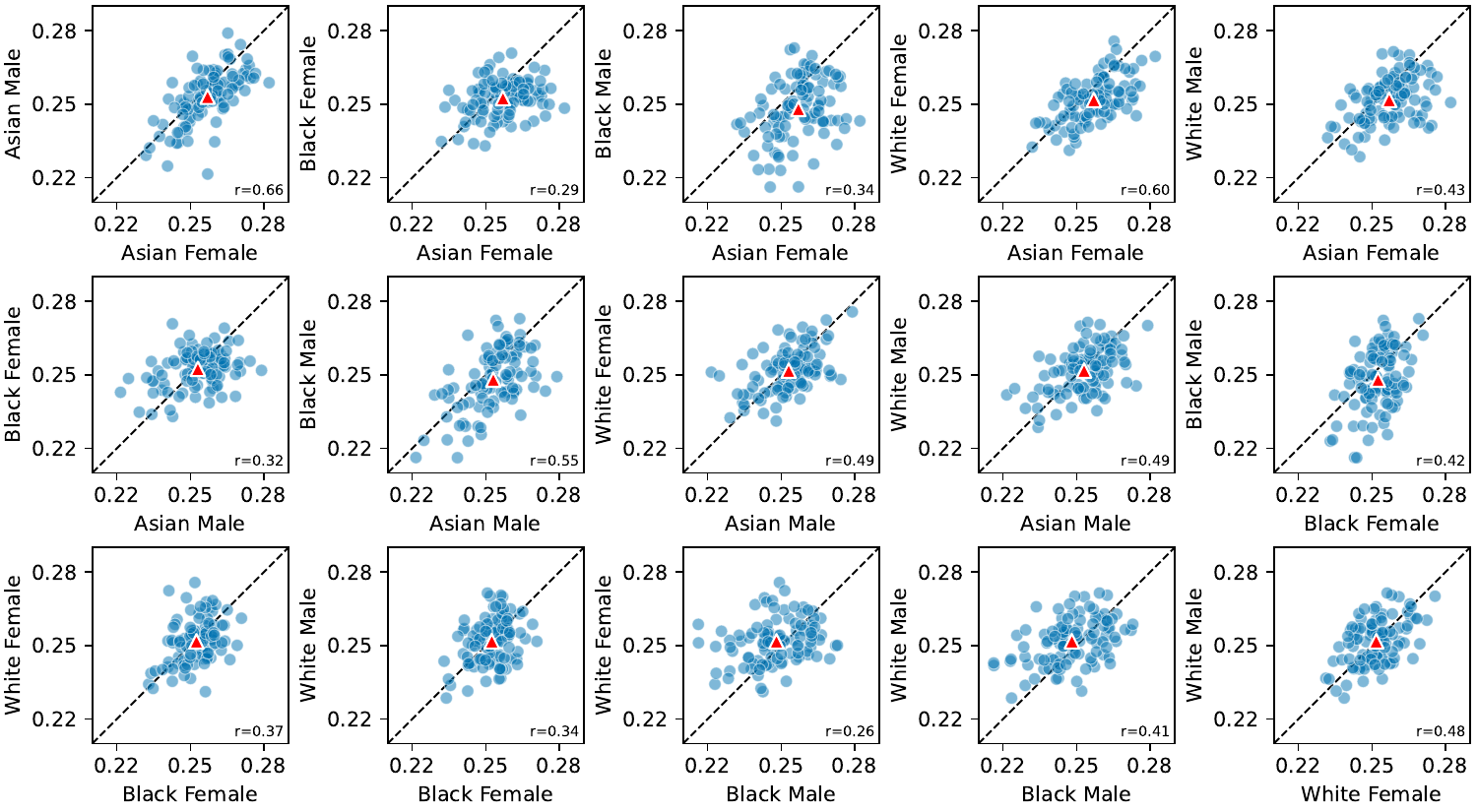}
    \caption{\textbf{Scatter plots showing cosine similarities to a neutral text prompt (e.g., ``a photo of a person'').} A single data point corresponds to two images from the same seed identity and different demographics, as indicated on each plot. Red triangles indicate mean values. We generally observe low to moderate Pearson correlations ($0.26 \leq r \leq 0.66$), indicating a notable effect of the seed identity.
    We conclude that uncontrolled image attributes (such as background, hairstyle, or face geometries) influence the association with a neutral text prompt and, thus, likely also with a valenced text prompt.}
    \label{fig:neutral-scatter}
\end{figure}

\autoref{fig:kde-corrections} shows two additional effects: First, CLIP can classify faces by gender and race, matching human perception (right column, rows 1--4).
Second, there is a small bias in the cosine similarities that is specific to race and gender (first column, see also \autoref{fig:neutral-boxplot}, and \autoref{fig:neutral-scatter}). This bias is removed alongside the individual face image bias when the first column is subtracted from the second to obtain the $\Delta$ cosine similarity.

\end{appendices}
\end{document}